%% file: source_file.tex
\pdfoutput=1

\documentclass[preprint]{elsarticle}
\usepackage[english]{babel}
\usepackage{lineno,hyperref}
\usepackage{afterpage}
\modulolinenumbers[5]

\journal{arXiv}

\biboptions{authoryear}

\usepackage{tikz, pgfplots}
\usepgfplotslibrary{fillbetween}
\usepackage{tkz-graph}
\usepackage{lscape,longtable}
\usepackage[utf8x]{inputenc}

\usepackage{amsfonts}
\usepackage{epsfig}
\usepackage{verbatim}
\usepackage{algorithm2e}
\usepackage{pdflscape}
\usepackage{mathtools}
\usepackage{float}
\usepackage[normalem]{ulem}

\usepackage{caption, subcaption}
\usepackage[para]{footmisc}

\newcommand{\OPT}{\mathrm{OPT}}

\newcommand{\R}{\mathbb{R}}

\newcommand{\Orm}{\mathrm{O}}
\newcommand{\Omegarm}{\mathrm{\Omega}}

\newcommand{\RC}{\mathrm{RC}}

\newcommand{\sop}{\textrm{SP}}
\newcommand{\op}{\textrm{P}}
\newcommand{\alg}{\textrm{A}}
\newcommand{\aop}{\alg_\op}

\newcommand{\iffr}{\textrm{EvFW}}
\newcommand{\ls}{\textrm{LS}}
\newcommand{\brkga}{\textrm{BRKGA}}
\newcommand{\ssm}{\textrm{SSM}}

\newcommand{\mir}{i}
\newcommand{\misli}{l}
\newcommand{\misbi}{b}
\newcommand{\tgf}{t}

\newcommand{\gic}{\textrm{GIC}}
\newcommand{\lic}{\textrm{LIC}}

\newcommand{\F}{\textrm{F}}

\newcommand{\raw}{\rightarrow}

\newcommand{\rfsc}{\Re} 

\begin{document}

\begin{frontmatter}

	\title{Evolutionary framework for two-stage stochastic resource allocation problems}
	\author[UNIFEI]{Pedro H. D. B. Hokama\fnref{hokamaGrant}}
	\ead{hokama@unifei.edu.br}
	\fntext[hokamaGrant]{Supported~by~FAPESP~2016/11082-6;}
	\author[UFSCAR]{M\'{a}rio C. San Felice\corref{mycorrespondingauthor}\fnref{marioGrant}}
	\ead{felice@ufscar.br}
	\fntext[marioGrant]{FAPESP~2017/11382-2;}
	\author[UEMS]{Evandro C. Bracht\fnref{evandroGrant}}
	\ead{evandro@uems.br}
	\fntext[evandroGrant]{FAPESP~03/13815-0;}
	\author[UNICAMP]{F\'{a}bio L. Usberti\fnref{fabioGrant}}
	\ead{fusberti@ic.unicamp.br}
	\fntext[fabioGrant]{CNPq~307472/2015-9.}
	\address[UNIFEI]{Federal University of Itajubá - UNIFEI, Itajubá MG, Brazil.}
	\address[UFSCAR]{Federal University of S\~{a}o Carlos - UFSCar, S\~{a}o Carlos SP, Brazil.}
	\address[UEMS]{State University of Mato Grosso do Sul - UEMS, Dourados MS, Brazil.}
	\address[UNICAMP]{University of Campinas - UNICAMP, Campinas SP, Brazil.}
\begin{abstract}

Resource allocation problems are a family of problems in which resources must be selected to satisfy given demands. This paper focuses on the two-stage stochastic generalization of resource allocation problems where future demands are expressed in a finite number of possible scenarios. The goal is to select cost effective resources to be acquired in the present time (first stage), and to implement a complete solution for each scenario (second stage), while minimizing the total expected cost of the choices in both stages. 

We propose an evolutionary framework for solving general two-stage stochastic resource allocation problems. In each iteration of our framework, a local search algorithm selects resources to be acquired in the first stage. A genetic metaheuristic then completes the solutions for each scenario and relevant information is passed onto the next iteration, thereby supporting the acquisition of promising resources in the following first stage. Experimentation on numerous instances of the two-stage stochastic Steiner tree problem suggests that our evolutionary framework is powerful enough to address large instances of a wide variety of two-stage stochastic resource allocation problems.

\end{abstract}
\end{frontmatter}


\section{Introduction} \label{sec:intro}

Resource allocation problems arise when companies are faced with the decision of choosing resources to build an infrastructure at a minimum cost which meets some given demands and constraints. 
However, in the real world, there are many uncertainties concerning the costs and demands of future resources. 
Stochastic programming is a field of research that is concerned with the modeling of these uncertainties \citep{birge2011introduction,shapiro2014lectures}. 
A common approach is to rely on a scenario decomposition analysis. The uncertainty about the problem parameters is modeled by a limited number of subproblems (scenarios), weighted by their occurrence probability and containing a restricted representation of the problem uncertainties.
By obtaining the optimal solutions for each scenario, one could expect to find similarities and trends among the scenarios to come up with a solution that holds a good trade-off under all scenarios.

Two-stage stochastic programming is a mathematical framework to model stochastic problems using a scenario decomposition approach.
The first stage reflects decisions that should be made at the present time.
On the other hand, the second stage reflects the decisions that should be made at a future time, considering a set of possible scenarios, each giving a presumable realization of the uncertain data.
The goal is to find the minimum cost solution, which comprises the first stage cost and the second stage expected cost considering all scenarios.


Literature is plenty of operations research problems modeled as two-stage stochastic problems. In the following, related works based upon real-life case studies offer examples of two-stage stochastic problems with an underlying resource allocation problem.
\begin{description}
	\item[Supply chain network design:] \cite{KaraOnut2010} developed a revenue maximization model for a network design problem, faced by a waste-paper recycling industry, in which one must find locations for recycling centers and flows among a multi-facility environment. 
	The model deals with the management of a Closed-Loop Supply Chain (CLSC), i.e., it considers both the forward flow (new paper from manufacturers to customers) and the reverse flow (waste paper from customers back to manufacturer). 
	\cite{Badri20171} introduce a model for the design of a CLSC in which demand and return volumes of a product are stochastic. The objective is to maximize the \textit{economic value added} (EVA) of the supply chain, considering supply chain costs, sales growth, working capital and fixed assets during a given planning horizon. In the first-stage, the decisions are made with respect to the number and location of facilities in a three-echelon logistics network. In the second-stage, the flows and storage of products are determined to each scenario, from which the EVA can be obtained. 
	\item[Inventory management:] \cite{Cunha2017313} propose a model for inventory control, within a given planning horizon, of a single-item one-echelon supply chain, considering periodic review of the stock position and uncertainty of the demand levels from the retailer. The first-stage decisions concern the optimization of two inventory control variables: (\textit{i}) the review periodicity of the stock level and (\textit{ii}) the target level for each stock replenishment order. The second-stage decisions refer to the stock levels and to the quantities ordered over the periods of the planning horizon, which are directly influenced by the first-stage decisions and the realizations of the uncertain demands. The objective is to minimize the costs of ordering, carrying and shortage.
	\item[Airline operations:] \cite{McCarty20181} propose a model for the decision-making of anticipating passenger reaccommodation from delayed flights. The first-stage decisions assign passengers to new itineraries in anticipation of the delay's impact and second-stage decisions adjust itineraries for passengers who missed their connections once the delay has been realized. \cite{Carreira2017639} investigate an airline fleet planning problem with the objective to select aircrafts to purchase (first-stage) or lease (second-stage) in order to satisfy all passengers demands.
	\item[Disaster relief:] \cite{Krasko2017265} present an optimization model to manage hazardous post-fire debris flow and to coordinate pre-disaster mitigation (first-stage) with disaster response (second-stage). More specifically, in the first stage the decisions concern on mitigation actions (mulching, check dams, straw wattles and debris basin) and setting the number of vehicles to stock in each hospital. The second stage is composed of storm scenarios, each one deriving in a rescue vehicle routing problem. The recourse decisions are made within a multi-period framework aiming for the evacuation plan which minimizes casualties.
	\item[Tour scheduling:] \cite{Restrepo2017620} investigate a multi-activity tour scheduling problem with uncertain demands. A model is proposed to generate, in the first stage, weekly tours (days-off, working days, shift start times and shift lengths) for employees with identical skills (first-stage). In the second-stage a set of demand scenarios are given and for each scenario the working tours must cope with the realized demand. This is made by allocating work activities and breaks to the employees daily shifts while minimizing undercovering and overcovering of demand.
	\item[Agriculture:] \cite{Cobuloglu2017251} present a model for food and biofuel production incorporating economic and environmental impacts under yield and price level uncertainties. Sales revenue and the costs of seeding, production, harvesting and transportation at the farm level are considered as economic variables. The model also considers environmental effects including carbon emission and sequestration, soil erosion  and nitrogen leakage to water. The first-stage decisions regards allocating different areas of land to food and energy crops, while the second-stage variables are recourse decisions related to harvesting, budget allocation and amounts of yield types.
\end{description}
The numerous real-life applications arising from different fields being modeled as two-stage stochastic problems, emphasize the importance of developing efficient methodologies to tackle these problems.

\cite{RockafellerWets1991} introduced a paradigm to solve two-stage stochastic problems, called \textit{progressive hedging}. This method was originally proposed for problems with
continuous variables only, preferably linear problems. In the first step, a set of optimal solutions is obtained for the scenarios. Then, in the second step, through an averaging computation over all scenarios solutions, a good compromise first stage solution is built. \citeauthor{RockafellerWets1991} have shown that, by iterating the first and second steps, the progressive hedging converges to the optimum if the problem is convex.

\cite{LokketangenWoodruff1996} extend the ideas of \citeauthor{RockafellerWets1991} introducing an integer progressive hedging framework applicable to mixed-integer two-stage stochastic problems. Later, \cite{Watson2010} noticed that some issues arise when using this integer progressive hedging, especially for large-scale instances, resulting in either non-convergence or unacceptably long running-times. Moreover, \cite{Watson2010} proposed a number of algorithmic enhancements to improve the progressive hedging performance as a heuristic for stochastic mixed-integer programs. 

A two-stage mixed-integer stochastic problem can be reduced to a single mixed-integer programming model (MILP), often called \textit{deterministic equivalent} formulation. For practical problems, this extended formulation takes the form of a large scale MILP, since  all first and second stage variables, including all scenarios, are dealt with simultaneously.
For relatively small instances, mixed-integer programming solvers can be used to solve the deterministic equivalent model of the problem, as suggested by \cite{ParijaAhmedKing2004}. However, for practical resource allocation problems, the deterministic equivalent forms are usually too large to tackle, even with a state-of-the-art MIP solver.
Moreover, \cite{TometzkiEngell2009} points out that MILP solvers still do not exploit the staircase structure
of these deterministic formulations.

To avoid the computational burden of solving a two-stage stochastic problem through its deterministic MILP model, \cite{TillEtAl2007} proposed a stage decomposition approach. Their method uses an evolutionary metaheuristic to explore and optimize the first stage decision variables. For each set of values explored by the metaheuristic for the first stage variables, the full second stage recourse cost is calculated for the decoupled scenarios. This is made by applying a MILP solver to each scenario independently. Both papers perform computational experiments on real-world scheduling problems with uncertain demands and capacities. The results show that the stage decomposition approach delivered better solutions than by solving the deterministic model, within limited computational time.
Later, \cite{TometzkiEngell2011} showed that the stage decomposition approach proposed by \cite{TillEtAl2007} can potentially benefit from good starting solutions for the evolutionary algorithm. Numerical experiments show that initialization methods using mathematical programs can significantly improve the results compared to random initialization of the evolutionary algorithm population.

\cite{AmorimCostaAlmada2015} presented a hybrid method to build algorithms, that combine a mixed-integer linear solver with a path-relinking metaheuristic, to solve two-stage stochastic problems with continuous second stage
decision variables. 
In the first step of this hybrid algorithm, each scenario is solved individually by a mixed-integer linear solver. The solutions obtained for each scenario 
are ranked by their stochastic costs.
The best ones are used by the path-relinking phase as guiding solutions to better explore the solution space. 
The hybrid method was tested considering a stochastic lot sizing and scheduling problem, and results have shown that the method outperformed the use of the mixed-integer linear solver alone, especially for the hardest instances.

In the past few decades, there have been staggering increases on the rate by which new data sets are generated, as well as the amount of stored data collected by numerous devices (mobile phones, software logs, wireless sensor networks). This phenomena is commonly referred to as \textit{big data} \cite{Han-etal-2014}, which offers modern resource allocation problems the access to a continuous flow of information. Therefore, efficient optimization methods, capable of dealing with large volumes of data, are in need. This matter is even more relevant for two-stage stochastic problems, since an instance size is also affected by the number of scenarios.

\subsection*{Our contributions}


This work proposes an efficient heuristic methodology, called \textit{evolutionary framework} (\iffr), to tackle large-scale general two-stage stochastic optimization problems.
It contains a two-step main loop, each step responsible for solving one stage and supplied with relevant information from the previous step. Briefly, the first step selects first stage resources until a local optimum is attained. The second step solves each second stage scenario using a \textit{biased random keys genetic algorithm} (BRKGA).
Our proposed methodology, to the best of our knowledge, is the first fully heuristic stage decomposition approach to solve two-stage stochastic resource allocation problems. The fact that \iffr\ does not rely on ILP solvers allows it to tackle larger instances, that derive from modern resource allocation problems.

The two-stage stochastic Steiner tree problem (SSTP) was selected to showcase the \iffr\ effectiveness. 
The SSTP is an NP-hard network design problem with many real-life applications and it was recently selected to compose the 11th DIMACS Implementation Challenge~\citep{Dimacs2014}. 
Computational tests were performed on 24 instances from \cite{Bomze2010}, and 560 much larger instances from DIMACS~\citep{Dimacs2014} and first solved by us in~\cite{Hokama2014}.
For the former, our methodology obtains near-optimal solutions in short computational times.
For the latter, within reasonable computational times, it obtains cost-effective solutions for instances with 20 times more scenarios than those from the literature. 
These results produce a new benchmark for stochastic optimization methods to solve large-scale resource allocation problems.
To the best of our knowledge, these are all the instances with results reported in the literature.

This paper is organized as follows. 
Section~\ref{sec:Rap} provides a brief introduction to (non-stochastical) resource allocation problems, accompanied with their mathematical formulations. This section also gives the basic concepts of the BRKGA metaheuristic and how it can be applied to solve these problems. Section~\ref{sec:two-stage} extends the mathematical formulations given in the previous section for general two-stage stochastic resource allocation problems, which are the focus of this work. Section~\ref{sec:Framework} presents a detailed description of our proposed methodology, \iffr, addressing the first stage local search, second stage metaheuristic, feedback between iterations, and convergence criteria. Section~\ref{sec:case-study} shows a case study on the application of \iffr\ for the two-stage stochastic Steiner tree problem. Computational experiments for this case study are performed and discussed in the same section, comparing \iffr\ with results from the literature. Finally, Section~\ref{sec:conclusion} presents our considerations and final remarks.

\section{Resource allocation problems} \label{sec:Rap}

We consider a generic (non-stochastic) resource allocation problem, which will be denoted by \op, where resources with non-negative costs are selected to build a minimum cost infrastructure that meets given demands and some side constraints.
Side constraints restrict the possible combination of selected resources and must be respected in any feasible solution.

Let $R$ be a resource set, $c: R \rightarrow \mathbb{R}^+$ be a cost function, and $D$ be a demand set.
We define $I = (R, c, D)$ as an instance for \op, and $\rfsc$ as the family of all combinations of resources that respect the  given side constraints of \op.
Let $f$ be a function that maps $D$ to the sub-family $f(D) \subseteq \rfsc$ of all resource sets that attend $D$.

A solution for instance $I = (R, c, D)$ of problem \op\ is a set $R' \in \rfsc$ such that $R'$ belongs to $f(D)$, that is, the set of resources $R'$ attends $D$ and respects the side constraints.
The goal is to find a solution $R'$ with a minimum cost. 
Therefore, problem \op\ can be formulated as
\begin{align}
 \label{eq:OP_sol_cost}
 \min & \sum_{r \in R'} c(r) \enspace,\\
 \mbox{subject to} \quad  & R' \in (\rfsc \cap f(D)) \nonumber \enspace.
\end{align}

Now, let $\aop$ be a generic algorithm for a generic problem \op, that receives an instance $I = (R, c, D)$ and returns a solution $R'$. We consider that $\aop$ is any simple and fast heuristic for \op, such that it does not return an optimal solution.
In the following subsections, we briefly describe the metaheuristic BRKGA and how to combine it with algorithm $\aop$ to find good solutions for~\op.

\subsection{BRKGA overview}

The Biased Random Key Genetic Algorithm (BRKGA), presented by \cite{Goncalves2011}, is a general search metaheuristic based on genetic algorithms, where a population of individuals evolves through the Darwinian principle of the survival of the fittest.

Each individual $j$ of the population is represented by a chromosome $Q^j$, encoded as a vector with $m$ alleles.
Each allele is a random key uniformly drawn over the interval $[0, 1]$.
The decoder is an algorithm that translates a chromosome $Q^j$ into a solution $R^j$.
The fitness is a function that evaluates a solution $R^j$.

The BRKGA initializes the population with $p$ randomly generated chromosomes, and this population evolves along $g$ generations.
For $i = 1, \dots, g-1$, the population is updated from generation $i$ to $i+1$ according to the following steps:

\begin{enumerate}
 
 \item For each individual $j$ at $i$, decode the chromosome $Q^j$ into a solution $R^j$.
 
 \item Evaluate the fitness of each solution $R^j$ for each individual $j$ at $i$.
 
 \item Copy the best $p_e$ individuals (elite set) from $i$ to $i+1$.
 
 \item Add $p_m$ randomly generated chromosomes (mutants) to $i+1$.
 
 \item Produce $p - (p_e + p_m)$ new chromosomes to $i+1$ using crossovers.
 
\end{enumerate}

The crossover generates a new individual by sampling each allele from one of its parents.
Both parents are from the current generation and exactly one is from the elite set. 
An allele is sampled from the elite parent with probability $\rho_e$.

After $g$ generations, the BRKGA returns the best individual's chromosome $Q^*$ and its decoded solution $R^*$.

\subsection{BRKGA and resource allocation problems} \label{sec:BRKGA_op}

We show a method to apply the BRKGA to a resource allocation problem \op.
This method is particularly interesting for dealing with NP-hard problems for which some fast approximation algorithm or heuristic is known.

Considering an instance $I = (R, c, D)$ and a BRKGA individual $j$, we define the following terms:

\paragraph{Chromosome}

Each allele of chromosome $Q^j$ corresponds to a resource $r \in R$ with key $Q^j(r)$. 
Thus, the chromosome $Q^j$ is a vector of size $|R|$.

\paragraph{Initial population}

We replace exactly one randomly generated chromosome from standard BRKGA initial population by a regular chromosome $Q$, in which all allele values are equal to $0.5$, this will be a neutral chromosome.

\paragraph{Decoder}

For each resource $r \in R$ and chromosome $j$, the allele key $Q^j(r)$ is used as a perturbation for the corresponding resource cost $c(r)$. 
Let $\alpha \in {\mathbb R}^+_*$ be a parameter that determines the perturbation intensity. 
The new perturbed cost of $r$ is
\begin{equation} \label{eq:alt_costs}
 c^j(r) = (1 - \alpha + 2 \,\alpha \,Q^j(r))\, c(r) \enspace.
\end{equation}
Now, we decode $Q^j$ into a solution $R^j$ by using $\aop$ to solve a new instance $I^j = (R, c^j, D)$.

Note that for any $r \in R$, $Q^j(r) < 0.5$ gives a discount to the cost of resource $r$, while $Q^j(r) > 0.5$ increases its cost. 
The idea is that the perturbed costs allow $\aop$ to search over, possible good, solutions that were not considered when $\aop$ worked with the original costs.
Moreover, the regular chromosome added to the initial population guarantees that the original costs are also considered by $\aop$.

\paragraph{Fitness}

Individual $j$ fitness is the cost of the decoded (from $Q^j$) solution $R^j$ evaluated with the original costs, i.e. $\sum_{r \in R^j} c(r)$. It is important to note that perturbed costs are passed to $\aop$ to find $R'$ but are not considered to evaluate the real cost of $R'$.

We denote by {\tt BRKGA($I, \aop, Q$)} a call for \brkga\ to solve input $I$ of \op, using algorithm $\aop$, and adding a chromosome $Q$ to the initial population.

\section{Two-stage stochastic resource allocation problems} \label{sec:two-stage}

We now describe the family of two-stage stochastic resource allocation problems, based on (non-stochastic) resource allocation problems defined in the previous section. Our framework \iffr\ aims to solve this family of problems. 

Formally, given a resource allocation problem \op, we denote as \sop, the two-stage stochastic version of \op. We define \sop\ as follows:
Let $S$ be a scenario set, such that each scenario $s \in S$ has probability $p_s$ to occur.
Let $R$ be a resource set, such that each resource $r \in R$ has first stage cost $c_0(r)$ and, for each scenario $s \in S$, resource $r$ has a second stage cost $c_s(r)$.
Furthermore, let $D_s$ be the demand set to be attended for each scenario $s \in S$.
 
Considering probabilities $Pr = \{p_s : s \in S\}$, cost functions $C = \{c_0, c_s : s \in S\}$, and demands ${\cal D} = \{D_s : s \in S \}$, we define ${\cal I} = (S, Pr, R, C, {\cal D})$ as an instance for \sop.
A solution for \sop\ is defined by a set of resources to be acquired in the first stage $R_0 \subseteq R$ and, for each $s \in S$, a set of resources $R_s \subseteq R \backslash R_0$, such that $R_0 \cup R_s$ is a solution for instance $I_s = (R, c_s, D_s)$ of \op. Note that it is necessary to find a solution for each possible scenario.
The goal is to find an \sop\ solution with the minimum cost.
Therefore, 
the \sop\ can be formulated as
\begin{align}
 \label{eq:SOP_sol_cost}
 \min &  \sum_{r \in R_0} c_0(r) + \sum_{s \in S} \sum_{r \in R_s} c_s(r) \cdot p_s \enspace, \\ 
 \mbox{subject to} \quad  & (R_0 \cup R_s) \in (\rfsc \cap f(D_s)), \quad \forall s \in S \nonumber  \enspace.
\end{align}

Note that the two-stage stochastic versions of relevant problems from combinatorial optimization and operations research, such as set cover \citep{ChristofidesKorman1975}, network design \citep{JohnsonLenstraKanRinnooy1978}, facility location \citep{Drezner1995}, and vehicle routing \citep{Caceres-Cruz:2014} problems, fit into this generic problem definition.

For an instance ${\cal I}$ of \sop, the recourse cost function $\F(R_0, \aop, {\cal I})$ is the cost of solving each scenario of ${\cal I}$ with $\aop$, considering cost zero for first stage resources $R_0$. Intuitively this is the cost of completing the solution for each scenario, given that resources $R_0$ are 
already acquired and paid in the first stage.
More formally, for each scenario $s \in S$, we define 
\begin{equation}
 c'_s(r) = \left\{ \begin{array}{ll} 
  0 & \mbox{if $r \in R_0$,}\\
  c_s(r) & \mbox{otherwise,}
 \end{array} \right.
\end{equation}
and let $R_s$ be the solution obtained by algorithm $\aop$ when it solves instance $I_s = (R, c'_s, D_s)$ of \op, for each scenario $s$ of instance ${\cal I}$ of problem \sop.
We define the recourse cost function  
\begin{equation} \label{eq:rec_cost_func}
\F(R_0, \aop, {\cal I}) = \sum_{s \in S} \sum_{r \in R_s} c_s(r) \cdot p_s \enspace.
\end{equation}

It is worth mentioning that an upper bound for the optimum cost of \sop\ consists of letting all resources to be acquired at each scenario of the second stage using the $\aop$, i.e., no resource is acquired in first stage and $R_0 = \emptyset$. Moreover, note that if $\gamma$ is the highest inflation ratio of the resources and $\aop$ is a $\beta$-approximation algorithm for \op, then this upper bound is a $\beta (1+\gamma)$-approximation for the optimum cost of \sop.

Formally, let $\F(\emptyset, \aop, {\cal I})$ be the value of this upper bound, $\OPT({\cal I})$ be the cost of an optimal solution, and $\OPT_{\op} (I_s)$ be the cost of an optimal solution for instance $I_s = (R, c_s, D_s)$ of \op.
We have that 
\[
\OPT({\cal I}) \quad\leq\quad \F(\emptyset, \aop, {\cal I}) \quad\leq\quad \beta \, \sum_{s \in S} p_s \OPT_{\op} (I_s) \quad\leq\quad \beta (1 + \gamma) \OPT({\cal I}) \enspace.
\]
If $\aop$ is an exact algorithm, the value of this upper bound is a $(1 + \gamma)$ approximation factor, that is, $
\OPT({\cal I}) \leq \F(\emptyset, \aop, {\cal I}) \leq (1 + \gamma) \OPT({\cal I}) \enspace.
$

\section{Evolutionary framework} \label{sec:Framework}

In this section, we describe the proposed Evolutionary Framework (\iffr) for the family of two-stage stochastic resource allocation problems. 
We first give an overview of the main algorithm, then we give details of each step, and finally we present a flow diagram, in Figure~\ref{fig:Flow_Diagram_EvFW}, summarizing the framework.

\paragraph{Overview}
The proposed evolutionary framework (\iffr) for an \sop\ has a main loop with two steps. 
In the first step, the local search algorithm (\ls) selects first stage resources towards a local optimum.
In the second step, the selected first stage resources are sent to the second stage metaheuristic (\ssm) that uses \brkga\ to solve each scenario of the second stage.
At the end of each \iffr\ iteration, a solution is obtained and relevant information is given as feedback for the next \iffr\ iteration.
After any convergence criterion is met, the main loop terminates and a final tail step uses the \ssm\ to refine the solution.
Algorithm~\ref{algo_iff} shows the pseudo-code of \iffr.
The subroutines are explained in the following subsections.

\IncMargin{1em}
\begin{algorithm}
  \SetKwFunction{EvFW}{EvolutionaryFramework}
  \SetKwFunction{LS}{LocalSearch}
  \SetKwFunction{SSM}{SecondStageMetaheur}
  \SetKwFunction{BRKGA}{BRKGA}
  \SetKwFunction{TS}{TailStep}
  \SetKwProg{myalg}{Algorithm}{}{}
  \SetKwData{Left}{left}\SetKwData{This}{this}\SetKwData{Up}{up}
  \SetKwData{And}{\textbf{and}}
  \SetKwFunction{StoppingCriteria}{StoppingCriteria}
  \SetKwFunction{Union}{Union}\SetKwFunction{FindCompress}{FindCompress}
  \SetKwInOut{Input}{input}\SetKwInOut{Output}{output}

  \myalg{\EvFW{${\cal I}$}}{
  \Input{${\cal I} = (S, Pr, R, C, {\cal D})$}
  \Output{$R_0$, ${\cal R} = \{ R_s : s \in S \}$}

  \BlankLine
  $R_0 \leftarrow \emptyset$ \\
  \For{$s \in S$ \And $r \in R$}{
    $Q_s(r) \leftarrow 0.5$
  }
  ${\cal Q} = \{ Q_s : s \in S \}$

  \BlankLine
  \While{stopping criteria are not satisfied}{
    $R_0 \leftarrow \LS({\cal I}, {\cal Q}, \aop)$ \\
    $({\cal Q}, {\cal R}) \leftarrow \SSM({\cal I}, {\cal Q}, \aop, R_0)$
  }
  ${\cal R} \leftarrow \TS({\cal I}, {\cal Q}, \aop, R_0)$
  \BlankLine
  }
  \caption{Evolutionary framework (\iffr).}\label{algo_iff}
\end{algorithm}\DecMargin{1em}

\subsection{Local search}

We first explain the local search algorithm (\ls) for the first \iffr\ iteration, which does not consider feedback information.
In the subsequent iterations, the \ls\ considers feedback from the \ssm, which we explain in Section~\ref{sec:Feedback}.

The \ls\ is responsible for identifying profitable resources to acquire in the first stage.
A resource is considered profitable if acquiring it does not increase the overall cost of the solution, i.e. its reduced cost (RC) is non-positive.
Let ${\cal I}$ be an instance for \sop\ and $R_0 \subseteq R$ be a set of resources already acquired in the first stage,
then we define the reduced cost of a resource $r$ in $R \backslash R_0$ as 
\[
 \RC(r, R_0, \aop, {\cal I}) = c_0 (r) + \F(R_0 \cup \{r\}, \aop, {\cal I}) - \F(R_0, \aop, {\cal I}) \enspace.
\]
Note that if $\aop$ is not an exact algorithm then the reduced cost is just an estimate of the resource actual profit.

The \ls\ begins with $R_0 = \emptyset$ and considers one resource at a time in an arbitrary order. 
For each resource $r$, the algorithm computes its reduced cost $\RC(r, R_0, \aop, {\cal I})$.
If $\RC(r, R_0, \aop, {\cal I}) \leq 0$ then $r$ is added to the first stage solution, i.e. $R_0 \leftarrow R_0 \cup \{r\}$. 
Note that this algorithm uses a first improvement approach.

Since the addition of new resources to $R_0$ may change the reduced cost of resources already in $R_0$, the \ls\ verifies if removing some of the resources previously acquired reduces the cost of the solution.
Considering each resource $r$ in $R_0$ in the order in which the resources were acquired, if 
\[
 \F(R_0 \backslash \{ r \}, \aop, {\cal I}) - \F(R_0, \aop, {\cal I}) - c_0 (r) < 0
\] 
then the algorithm removes $r$ from $R_0$.
The result is a set $R_0$ of resources to be acquired in the first stage.

The pseudo-code of the local search will be presented in Section \ref{sec:Feedback} when feedback information will be described.

\subsection{Second stage metaheuristic}

The second stage metaheuristic (\ssm) is responsible for solving each scenario, considering the resources acquired in the first stage.
Additionally, the \ssm\ uses the solution for each scenario to compose an \sop\ solution and  give feedback about promising resources to be acquired in the first stage (in the next iteration). Note that $R_0$ is the set of resources acquired in the first stage by the \ls.

For each scenario $s$, the \ssm\ consider an instance $I_s = (R, c'_s, D_s)$ of \op\ where $c'_s(r) = 0$ if a resource $r$ is in $R_0$, and $c'_s(r) = c_s(r)$ otherwise. Then we solve instance $I_s$ with BRKGA and $\aop$, as shown in Section~\ref{sec:BRKGA_op}, obtaining chromosome $Q^*_s$ with solution $R^*_s$ for the best individual. The cost for the \sop\ solution can be obtained by equation~(\ref{eq:SOP_sol_cost}).

The \ssm\ receives the set $\cal Q$ of best chromosomes ${\cal Q}_s$ for each scenario $s$ from a previous iteration. In the first iteration, all alleles of all chromosomes are set to 0.5. Obviously \ssm\ returns the set ${{\cal Q}^*} = \{Q^*_s : s \in S\}$ of best chromosomes found in the current iteration as feedback for the next \iffr\ iteration. Algorithm \ref{algo_ssm} presents the pseudo-code for this procedure.

\IncMargin{1em}
\begin{algorithm}
  \SetAlgoLined\DontPrintSemicolon
  \SetKwFunction{LS}{LocalSearch}
  \SetKwFunction{SSM}{SecondStageMetaheur}
  \SetKwFunction{BRKGA}{BRKGA}
  \SetKwFunction{TS}{TailStep}
  \SetKwProg{myalg}{Algorithm}{}{}
  \SetKwData{Left}{left}\SetKwData{This}{this}\SetKwData{Up}{up}
  \SetKwData{And}{\textbf{and}}
  \SetKwFunction{StoppingCriteria}{StoppingCriteria}
  \SetKwFunction{Union}{Union}\SetKwFunction{FindCompress}{FindCompress}
  \SetKwInOut{Input}{input}
  \SetKwInOut{Output}{output}
  
  \myalg{\SSM{${\cal I}, {\cal Q}, \aop, R_0$}}{
    \Input{${\cal I} = (S, Pr, R, C, {\cal D})$, ${\cal Q} = \{ Q_s : s \in S \}$, $\aop$, $R_0$}
    \Output{${\cal Q^*} = \{ Q^*_s : s \in S \}$, ${\cal R^*} = \{ R^*_s : s \in S \}$}

    \BlankLine
    \For{$s \in S$}{
      \For{$r \in R$}{
        \lIf{$r \in R_0$}{$c'_s(r) \leftarrow 0$}
        \lElse{$c'_s(r) \leftarrow c_s(r)$}  
      }
      $I_s \leftarrow (R, c'_s, D_s)$ \\
      $(Q^*_s, R^*_s) \leftarrow \BRKGA(I_s, \aop, Q_s)$
    }
  }
  \caption{Second stage metaheuristic (SSM).}\label{algo_ssm}
\end{algorithm}\DecMargin{1em}

\subsection{Feedback} \label{sec:Feedback}

From the second \iffr\ iteration onwards, similar to the \ssm, the \ls\ receives as feedback the chromosome set ${\cal Q} = \{Q_s : s \in S\}$, where each $Q_s$ is the chromosome for the best individual in scenario $s$ from the previous \iffr\ iteration.

We define the weighted average allele, for each resource $r \in R$, as 
\begin{equation} \label{eq:average_key}
 \bar{Q}(r) = \sum_{s \in S} Q_s(r) \cdot p_s \enspace.
\end{equation}
Recall that in the first \iffr\ iteration, the \ls\ begins with $R_0 = \emptyset$ and considers one resource at a time, to be added to $R_0$, in an arbitrary order. 
From the second \iffr\ iteration onwards the \ls\ uses a non-increasing order over $\bar{Q}$.
This is motivated by the notion that resources with low weighted average allele were likely used in the best solution for several scenarios in the previous \iffr\ iteration.
Thus, these resources represent promising acquisitions for the first stage.

For each scenario $s \in S$, using perturbed costs from~\eqref{eq:alt_costs}, we define 
\begin{equation} \label{eq:c'_s}
 c'_s(r) = \left\{ \begin{array}{ll} 
  0 & \mbox{if $r \in R_0$,}\\
  (1 - \alpha + 2 \alpha Q_s(r)) c_s(r) & \mbox{otherwise.}
 \end{array} \right. 
\end{equation}
Let $R_s$ be the solution obtained by $\aop$ when it solves instance $I_s = (R, c'_s, D_s)$ of \op.
We redefine the recourse cost function~\eqref{eq:rec_cost_func} as 
\begin{equation} \label{eq:F}
 \F(R_0, \aop, {\cal I}, {\cal Q}) = \sum_{s \in S} \sum_{r \in R_s} c_s(r) \cdot p_s \enspace.
\end{equation}
Note that while the costs $c'_s$ from~\eqref{eq:c'_s} bias $\aop$ when building solution $R_s$ for scenario $s$, these costs are not used at (\ref{eq:F}).
Using this new recourse cost function, we redefine the reduced cost as
\begin{align}
 \RC(r, R_0, \aop, {\cal I}, {\cal Q}) &= c_0 (r) + \F(R_0 \cup \{r\}, \aop, {\cal I}, {\cal Q}) \nonumber\\
 &~~~ - \F(R_0, \aop, {\cal I}, {\cal Q}) \enspace.
\end{align}

By using the redefined $\F$ and $\RC$ functions, the \ls\ bias the construction of second stage solutions to take advantage of resources with low weighted average allele.
Observe that both the new recourse cost function and the new reduced cost function are equivalent to the original ones if for every $Q \in {\cal Q}$ and $r \in R$ we have $Q(r) = 0.5$. Algorithm \ref{algo_ls} presents the pseudo-code of the Local Search Algorithm.

\IncMargin{1em}
\begin{algorithm}[H]
  \SetAlgoLined\DontPrintSemicolon
  \SetKwFunction{LS}{LocalSearch}
  \SetKwFunction{SSM}{SecondStageMetaheur}
  \SetKwFunction{BRKGA}{BRKGA}
  \SetKwFunction{TS}{TailStep}
  \SetKwProg{myalg}{Algorithm}{}{}
  \SetKwData{Left}{left}\SetKwData{This}{this}\SetKwData{Up}{up}
  \SetKwData{And}{\textbf{and}}
  \SetKwFunction{StoppingCriteria}{StoppingCriteria}
  \SetKwFunction{Union}{Union}\SetKwFunction{FindCompress}{FindCompress}
  \SetKwInOut{Input}{input}
  \SetKwInOut{Output}{output}

  \myalg{\LS{${\cal I}, {\cal Q}, \aop$}}{
    \Input{${\cal I} = (S, Pr, R, C, {\cal D})$, ${\cal Q} = \{Q_s : s \in S\}$, $\aop$}
    \Output{$R_0$}

    \BlankLine
    $R_0 \leftarrow \emptyset$ \\
    \For{$r \in R$}{
      $\bar{Q}(r) \leftarrow \displaystyle \sum_{s \in S} Q_s(r) \cdot p_s$
    }
    \For{$r \in R$ in increasing order of $\bar{Q}(r)$}{\label{for:ls}
      \lIf{$c_0 (r) + \F(R_0 \cup \{r\}, \aop, {\cal I}, {\cal Q}) - \F(R_0, \aop, {\cal I}, {\cal Q}) \leq 0$}{
        $R_0 \leftarrow R_0 \cup \{r\}$
      }
    }
    \For{$r \in R_0$ in order of acquisition}{
      \lIf{$\F(R_0 \setminus \{r\}, \aop, {\cal I}, {\cal Q}) - \F(R_0, \aop, {\cal I}, {\cal Q}) - c_0 (r) < 0$}{
        $R_0 \leftarrow R_0 \setminus \{r\}$
      }
    }
  }
  \caption{Local search (LS).}\label{algo_ls}
\end{algorithm}\DecMargin{1em}

\subsection{Convergence criteria and tail step}

At the end of each iteration of the \iffr\ main loop, some convergence criteria are checked to decide if the algorithm should keep searching for better solutions in new iterations, or just refine the best solution achieved so far in the tail step.

The two basic convergence criteria need the parameters $\mir \in {\mathbb R}^+_*$, $\misbi \in {\mathbb N}$ and $\misli \in {\mathbb N}$, that correspond to the minimum improvement ratio, the maximum number of \iffr\ iterations since the best solution was found, and the maximum number of \iffr\ iterations since the last solution improvement, respectively.

Let $S^*$ be the best solution found so far, $S'$ be the solution from the last \iffr\ iteration, and $S$ be the current solution.
If $c(S^*) > (1 + \mir) c(S)$ then the global improvement counter (\gic) is reset to $0$.
Otherwise, the \gic\ increases by $1$.
Similarly, if $c(S') > (1 + \mir) c(S)$ then the local improvement counter (\lic) is reset to $0$.
Otherwise, the \lic\ increases by $1$.
When the \gic\ equals $\misbi$ or the \lic\ equals $\misli$ then the respective convergence criterion is achieved.
The idea behind these convergence criteria is to allow, up to a certain limit, the algorithm to search for solutions out of a local minimum.
Another convergence criterion that may be used is a time limit.

If any of the convergence criteria is achieved, the \iffr\ main loop ends and the algorithm obtains the best solution $S^* = (R_0, {\cal R})$ so far, as well as the set of chromosomes ${\cal Q}$ associated with it, and uses a tail step to improve the solution achieved in each scenario.
We use {\tt TailStep(${\cal I}, {\cal Q}, \aop, R_0$)} to denote a call for the tail step algorithm.
This algorithm is identical to the SSM, except that in its calls to BRKGA the number of generations is multiplied by a tail generations factor~($\tgf$).

\input{flow_diagram_EvFW}

\section{Case study: stochastic Steiner tree} \label{sec:case-study}

In this section, we apply our \iffr\ to the two-stage stochastic Steiner tree problem (SSTP), and show some computational results. We also compare our results with those from the literature.

The Steiner tree problem in graphs (STP) is a classical network design problem.
Its goal is to find a minimum cost tree that spans a given subset of nodes, called terminals.
This combinatorial optimization problem has several applications, including: communication networks and power systems \citep{MagnantiWong1984}, wire routing in VLSI circuits \citep{Lengauer1990}, and the study of phylogenetic trees  \citep{CavalliEdwards1967}.
The STP supposes full knowledge of the terminals to be connected and of the edge costs involved.

The SSTP is a version of the STP that uses a set of possible scenarios to capture uncertainty, both from the terminal set to be connected, and from the costs of the edges.
Each scenario is characterized by a terminal set, second stage edge costs and a probability of occurrence.
In the SSTP, some edges are acquired in the first stage, considering the set of possible scenarios, and in each scenario of the second stage, some other edges are acquired, at an inflated cost, to complete a tree that connects the scenario's terminal set (in fact, a subgraph that contains such a tree).
The cost of a solution is the sum of first stage edge costs plus the expected second stage edge costs.

\input{SSTP_fig_1}
Figure~\ref{fig:1} depicts an instance of the SSTP with ten nodes and five scenarios.
The top leftmost graph shows the first stage edge costs, and the other graphs show, for each scenario, the terminal nodes, the second stage edge costs and the occurrence probability.

\input{SSTP_fig_2}
Figure~\ref{fig:2} depicts the optimal solution for the instance shown in Figure~\ref{fig:1}.
The top leftmost graph shows the edges acquired in the first stage, and the other graphs show, for each scenario, the edges acquired in the second stage.
Notice that, in each scenario, the union of first and second stage edges contains a tree that connects the scenario's terminal nodes.

\cite{Gupta07} investigated the SSTP, showing a $40$-approximation algorithm for the special case of the SSTP in which the second stage costs are determined by a fixed inflation ratio.
This algorithm is based on a primal-dual scheme, guided by a relaxed integer linear programming (ILP) solution.
\cite{Swamy2006} presented a $4$-approximation algorithm, that uses cost-sharing properties, for this SSTP fixed inflation ratio special case.
\cite{Gupta07} have also shown that the general case (non-fixed inflation ratio) is as hard as the label cover problem, whose approximation ratio lower bound is $\Omegarm(2^{\log^{1-\epsilon} n})$.

\cite{Bomze2010} proposed a two-stage branch-and-cut algorithm that consists of a semi-directed ILP model with integer L-shaped cuts, that is stronger than the undirected ILP model proposed by \citeauthor{Gupta07}.
\citeauthor{Bomze2010} were the first to report computational results for the SSTP, showing optimal solutions for instances up to $50$ scenarios and $274$ edges.

A BRKGA based heuristic, which we call MH, was proposed by \cite{Hokama2014} specifically for the SSTP, from which the first non-trivial solutions for instances from the 11th DIMACS Implementation Challenge~\citep{Dimacs2014} were obtained. 
To the best of our knowledge, this is the only study proposing a heuristic approach for the SSTP. 
Four major improvements of the \iffr\ with respect to the MH are described next:

\begin{enumerate}[(i)]

    \item The \iffr\ uses a local search algorithm to decide which resources are bought in the first stage and it only uses the BRKGA inside each scenario of the second stage metaheuristic, while the MH is a straightforward implementation of the BRKGA, which uses a single call, with arbitrary cut and perturbation values, to decide both which edges are bought in the first stage and in each scenario, respectively.
    Moreover, the MH uses a local search procedure, but just to generate one initial chromosome.
    \item The BRKGA is applied independently to each scenario which led to solutions with better quality in a shorter time, as it allowed \iffr\ to combine the solutions obtained in each scenario. 
    We highlight that the standard application of BRKGA involves representing the entire solution by a single chromosome. The decomposition was possible in this case due to the independence among scenarios.
    \item Using several short cycles, instead of having one long first step followed by one long second step, helps to verify progress more frequently. This allowed the \iffr\ to finish much earlier for most instances and also enabled it to spend more time on instances which could benefit from the additional computational effort.
    \item Using feedback from the second step solution to better inform the following first step decisions. \iffr\ uses this feedback to bias the first step in selecting ``shortcuts'' drawn by the previous second step. By ``shortcuts'' it should be understood a set of resources with consistently low chromosome keys, indicating that they are frequently used among the scenarios. This same mechanism allows the algorithm to avoid bad sets of resources that would otherwise be selected due to the greedy nature of the first step.
\end{enumerate}

\subsection{STP as a resource allocation problem} \label{sec:STP}

In this subsection, we formally describe the (non-stochastic) STP as a resource allocation problem, defined in Section~\ref{sec:Rap} as \op, and show a classical approximation algorithm for the STP.
Moreover, we give an example on how the \brkga\ perturbed costs, proposed in Section~\ref{sec:BRKGA_op}, help to improve solutions for the STP.

Let $G = (V, E)$ be a graph, with the edges in $E$ being the resources (as defined in Section~\ref{sec:Rap}); $c: E \raw \R^+$ be an edge cost function; and $D \subseteq V$ be a set of terminals, that corresponds to the demands.
We define $I = (G, c, D)$ as an instance for the STP.
A solution for this problem is a tree $T \subseteq E$ that spans $D$, and the goal is to find such a tree with the minimum cost.
Therefore, the STP can be formulated as
\begin{align}
 \label{eq:STP_sol_cost}
 \min & \sum_{e \in T} c(e) \enspace,\\
 \mbox{subject to} \quad & \mbox{$T$ is a tree that spans $D$} \nonumber  \enspace.
\end{align}
Notice that in this problem, the side constraints correspond to restricting solutions to those representing a tree over the graph.

Both our method to solve a resource allocation problem \op~(presented in Section~\ref{sec:BRKGA_op}) and our framework to solve the related stochastic problem \sop~(presented in Section~\ref{sec:Framework}) rely on an algorithm $\aop$ for \op.
For the STP, we use the minimum spanning tree $2$-approximation algorithm~\citep{Va03}, called MST-approx, as algorithm $\aop$. 
The MST-approx creates a complete graph $G' = \{D, E'\}$, where each edge $(i, j) \in E'$ has a cost equal to the shortest path between $i$ and $j$ in $G$.
Then, it computes a minimum spanning tree $T'$ of $G'$.
Finally, a Steiner tree $T$ for $G$ is derived from the union of the shortest paths represented by the edges of $T'$.
All steps of this algorithm can be performed in polynomial time.
More precisely, the time complexity to compute the shortest paths between all the terminals is $\Orm(|D| |E| \log |V|)$, the time complexity to compute a minimum spanning tree in $G'$ is $\Orm(|E'| \log |D|) = \Orm(|D|^2 \log |D|)$, and the time complexity to translate a minimum spanning tree in $G'$ to a tree in $G$ is $\Orm(|D||E|)$.
Thus, the MST-approx runs in $\Orm(|D| (|E| \log |V| + |D| \log |D| + |E|)) = \Orm(|D| |E| \log |V|)$.

\input{BRKGA_fig_1}
As described in Section~\ref{sec:BRKGA_op}, we use perturbed costs in the decoding process of \brkga\ to solve resource allocation problems. 
Figure~\ref{fig:decoder} illustrates the advantage of using these perturbed costs for the STP.
Consider the instance illustrated in Figure~\ref{fig:decoder_a}, in which the edge set is $E = \{ a, b, c, d, e, f \}$, the edge costs given by $c : E \raw \R^+$ are equal to $[9, 5, 9, 5, 5, 9]$, respectively, and the terminals are represented by red nodes.
Figure~\ref{fig:decoder_b} shows a solution with cost $18$, returned when the MST-approx considers the original edge costs.
Supposing a chromosome $Q = [0.7, 0.4, 0.5, 0.5, 0.3, 0.6]$, we have that the perturbed edge costs, computed by $c'(e) = (1 - \alpha + 2\alpha Q(e)) \cdot c(e)$ with $\alpha = 0.5$, are $[10.8, 4.5, 9, 5, 4 , 9.9]$.
When the MST-approx considers the perturbed edge costs, it returns an improved solution, represented by Figure~\ref{fig:decoder_c}, with original cost~$15$.

\subsection{SSTP as a two-stage stochastic resource allocation problem} \label{sec:SSTP}

In this subsection, we formally describe the SSTP as a two-stage stochastic resource allocation problem, defined in Section~\ref{sec:two-stage}.

Let ${\cal I} = (S, Pr, G, C, {\cal D})$ be an instance for the SSTP, where $S$ is a scenario set; $Pr = \{p_s : s \in S\}$, where $p_s$ is the occurrence probability for scenario $s$; $G = (V,E)$ is a graph; $C = \{c_0, c_s : s \in S\}$, where $c_0$ is the edge cost function for the first stage and $c_s$ is the edge cost function for scenario $s$; and ${\cal D} = \{D_s : s \in S \}$, where $D_s$ is the terminal set for scenario $s$.

A solution for the SSTP is defined by a set of edges $E_0 \subseteq E$ to be acquired in the first stage and, for each $s \in S$, a set of edges $E_s \subseteq E \setminus E_0$, such that $E_0 \cup E_s$ contains a tree that spans $D_s$.
The goal is to find a solution that minimizes the cost 
\[
\sum_{e \in E_0} c(e) + \sum_{s \in S} \sum_{e \in E_s} c_s(e) \cdot p_s \enspace.
\]

After reducing the STP to a resource allocation problem and the SSTP to a stochastic resource allocation problem, the application of the \iffr\ is straightforward.

\subsection{Computational results} \label{sec:results}

In this subsection, we show our computational results for the application of \iffr\ to the SSTP tested with two sets of instances.
The first set (benchmark~1), extracted from \cite{Bomze2010}, contains 24 instances with $5-50$ scenarios, $80-274$ edges and inflation ratio $\gamma \leq 0.5$.
The optimal solutions are known for all but one of these instances.
The second set (benchmark~2), extracted from the 11th DIMACS Implementation Challenge~\citep{Dimacs2014}, contains 560 instances, with $5-1000$ scenarios and $64-613$ edges\footnote{Instances available at: \url{ http://dimacs11.zib.de/downloads.html}\,.}.
These instances are divided into four types, \textit{K}, \textit{Lin}, \textit{P} and \textit{Wrp}, with inflation ratio~$\gamma$ bounded by $0.3$, $0.5$, $0.3$ and $1.0$, respectively.
Since our framework is non-deterministic, each instance tested in this section is executed with 20 different seeds (ranging from 1 to 20) and for each instance, we obtained the average and the standard deviation, both for cost values and time consumption.

Tables~\ref{tab:IFF_parameters} and~\ref{tab:BRKGA_parameters} show the default parameter settings for the \iffr\ and for the BRKGA, respectively.
These parameters were chosen after extensive computational tests.

\renewcommand{\arraystretch}{1.5}
\begin{table} [htb!]
\begin{center}
\caption{\iffr\ parameters.} \label{tab:IFF_parameters}
\begin{tabular}{r c l}
\hline
perturbation intensity & \hspace{0.5cm} & $\alpha = 0.7$ \\
minimum improvement ratio & \hspace{0.5cm} & $\mir = 0.001$ \\
max iterations since best solution & & $\misbi = 3$ \\
max iterations since last improvement & \hspace{0.5cm} & $\misli = 2$ \\
tail generations factor & & $\tgf = 3$ \\
\hline
\end{tabular}
\end{center}
\end{table}
\renewcommand{\arraystretch}{1}
\renewcommand{\arraystretch}{1.5}
\begin{table} [htb!]
\begin{center}
\caption{BRKGA parameters.} \label{tab:BRKGA_parameters}
\begin{tabular}{r c l}
\hline
population size & \hspace{0.5cm} & $\displaystyle p = 25$ \\
elite set size & & $p_e = \lceil 0.1p \rceil$ \\
number of mutants & & $p_m = \lceil 0.2p \rceil$ \\
elite crossover probability & & $\rho_e = 0.4$ \\
number of generations & \hspace{0.5cm} & $\displaystyle g = 25$ \\
\hline
\end{tabular}
\end{center}
\end{table}
\renewcommand{\arraystretch}{1}

Following we show some results for parameter tests which were executed over the smaller half of benchmark~2 instances (280 instances ranging from 5 to 150 scenarios).

First, we iteratively tested different combinations of parameters perturbation intensity ($\alpha$) and elite crossover probability ($\rho_e$), until these converged to $\alpha = 0.7$ and $\rho_e = 0.4$.
These are shown in Figures~\ref{fig:alpha} and~\ref{fig:rhoe}, respectively.
\input{graficos_alpha_rhoe}
%
The Improvement(\%) of each data point is the average improvement over all instances,
which 
is better explained during the upcoming analysis of benchmark~2. 
For all points in Figure~\ref{fig:param_local_convergence}, the average 
standard deviation over all instances is less than $0.19$ in the $y$-axis scale, which we omitted from the charts for clearance reasons.

We verified the behavior of \iffr\ applied to the SSTP when the population size~($p$) ranged over the values in $\{ 12, 25, 50, 100 \}$, and we show these results in Figure~\ref{fig:pop}.
\input{graficos_pop_blt}
%
We define a time ratio per pair (instance, population size) by dividing its average running time by the average running time for the same instance with population size $p = 25$.
The $x$-axis corresponds to the average time ratio over all instances.
We may see that while a larger population size leads to better results, the cost efficiency with respect to the time ratio decreases and the Improvement(\%) tends to asymptotically stabilize.

Moreover, we observed the behavior of \iffr\ when the parameters $\misbi$, $\misli$ and $\tgf$ (respectively, max iterations since best solution, max iterations since last improvement, and tail generations factor) range over the triples in $\{(0,0,0), (1,1,1), (3,2,3), (6,4,6), (10,6,10) \}$.
These results are shown in Figure~\ref{fig:blt} and their analysis is quite similar to the previous one.

For all points in Figure~\ref{fig:param_asymptotic_convergence}, the average standard deviation is less than $0.19$ in the $y$-axis scale and less than $0.18$ in the $x$-axis scale, which we omitted from the charts for clearance reasons.

Our computational experiments were executed on an Intel Xeon CPU E5 V3 2.30GHz with 57,6 GB RAM, under Ubuntu 16.04.
All codes were implemented in C++ using the Lemon Graph Library \citep{Lemon2014} as the framework for graph data structures.

\subsubsection*{Benchmark 1}

\cite{Bomze2010} reported the best solutions obtained by their branch-and-cut algorithms for benchmark~1.
In their paper, only one instance (lin05\_160\_269, $|S|=20$) remained without an optimal solution, given the time limit of two hours. 
Table~\ref{tab:Bomze} compares our results for benchmark~1 with those of exact methods Extensive Form ($EF$) and 2-stage Branch-and-Cut ($2BC^*$) from \cite{Bomze2010}.
For each instance, we report the average cost over the 20 tested seeds.
Since the instances for this benchmark are relatively small, we ran it with more permissive parameters than those from Tables~\ref{tab:IFF_parameters} and~\ref{tab:BRKGA_parameters}, i.e., $p = 100$, $b = 10$, $l = 6$ and $t = 10$.
The standard deviation for the solution costs is, on average, less than $0.5\%$ of the solution cost of each instance.

\input{tabela_bomze}

These results show that the \iffr\ presented a good performance for instances with known optimums, achieving small optimality gaps with an average of $1.17\%$. 
For the instance with unknown optimum, the \iffr\ has found a solution which is $2.3\%$ ($0.8\%$ on average) cheaper than the previous best known.
Moreover, the \iffr\ required an average of $197.1$ seconds per instance, while $EF$ required on average $1879.7$ seconds and $2BC^*$ required on average $1482.0$ seconds, which means an overall convergence significantly faster for our approach.
We highlight that these time comparisons are fair since the experiments from \cite{Bomze2010} used an Intel Core-i7 2.67GHz Quad Core machine with 12 GB RAM, whose single core speed is slightly faster than that from the Intel Xeon CPU E5 V3 2.30GHz used in our experiments.

\subsubsection*{Benchmark 2}

Table~\ref{tab:Dimacs} shows our summarized results for benchmark~2, displaying average values for each combination of instance type and number of scenarios\footnote{Detailed results at: \url{http://www.ic.unicamp.br/\textasciitilde fusberti/problems/sstp/}\,.}. 
Let \textit{buy\_none} be the upper bound $\F(\emptyset, \textrm{MST-approx}, {\cal I})$, described in Section~\ref{sec:two-stage}, in which no edge is acquired in the first stage and each scenario of the second stage is solved using the algorithm \textrm{MST-approx}.
Since we do not know the cost of the optimal solutions for the instances in this benchmark, we have used the \textit{buy\_none} upper bound to determine the relative cost reduction obtained by MH, the heuristic from \cite{Hokama2014}, and by the \iffr.
This comparison shows how much it is expected that each heuristic saves by anticipating future demands and acquiring cheaper edges in the first stage.
More precisely, $\delta c$~(\%) is the average of the improvement ratio achieved on each instance, which corresponds to one minus the ratio between the heuristic cost over the \textit{buy\_none} cost.
Therefore, higher $\delta c$ (\%) indicates better solutions.

\input{tabela_DIMACS}

For the smaller half of this benchmark, i.e., instances with $5$ to $150$ scenarios, we ran the tests with more permissive parameters than those from Tables~\ref{tab:IFF_parameters} and~\ref{tab:BRKGA_parameters}, i.e., $p = 100$, $b = 10$, $l = 6$ and $t = 10$.
For the larger half, we used the parameters from those tables.
Moreover, to promote a fair comparison among the execution times of MH and \iffr, we multiplied the MH times by a $1.6$ time ratio.
We ran an experiment with the smaller half of the instances on 
the same machine used for the original MH experiments 
and the $1.6$ was obtained as a lower bound for the resulting time ratio.

The results show that the \iffr\ is able to find non-trivial solutions for instances 
with number of scenarios and overall size well beyond what was possible for the exact branch-and-cut algorithms from \cite{Bomze2010}. 
We also observe that for sets of instances with 50 or more scenarios, in general, \iffr\ achieves higher improvements in significantly less computational time than the previous approach from \cite{Hokama2014}.
For the smaller half of the benchmark, the average cost reduction achieved by the \iffr, relative to the \textit{buy\_none} upper bound, was $4.01\%$ and took, on average, $1233.6$ seconds per instance.
For the larger half, the average cost reduction was $3.59\%$ and took, on average, $1397.9$ seconds per instance.
Furthermore, the average standard deviation of $\delta c$ (\%) is $0.17$ for the smaller half and $0.11$ for the larger half of benchmark 2.

We use Figure~\ref{fig:time_ratio_step} to analyze how the \iffr\ splits its processing time among its steps.
As shown in Figure~\ref{fig:ratio_pequenas}, when solving the smaller instances with more permissive parameters, i.e., $p = 100$, $b = 10$, $l = 6$ and $t = 10$, we have that \iffr\ spent $9\%$, $45\%$ and $46\%$ of its running time on first step, second step and tail step, respectively.
Similarly, Figure~\ref{fig:ratio_grandes} shows that when solving the larger instances with standard parameters, i.e., $p = 25$, $b = 3$, $l = 2$ and $t = 3$, \iffr\ spent $29\%$, $40\%$ and $31\%$ of its running time on each of those steps.
Notice that the first step corresponds to the Local Search (\ls) and that both the second and tail steps correspond to the Second Stage Metaheuristc (\ssm), albeit with a different number of generations.

We analyze that the different distribution of time among the framework steps in these experiments is due to the change in parameters $p$, $b$, $l$ and $t$. 
Notice that $p$ affects both the time spent on second and tail steps, $b$ and $l$ affects the time spent on first and second steps, and finally $t$ affects the time spent on tail step.
Thus, the time spent on second and tail steps is affected by a greater number of parameters.

\input{graficos_tempo_passos}

Furthermore, each inner cycle of Figure~\ref{fig:time_ratio_step} shows that \iffr\ spent more than $90\%$ of its time solving $\aop$, i.e., MST-approx.
This suggests that the \iffr\ is heavily dependent on an efficient $\aop$ algorithm.
However, for some problems it may be hard to find fast heuristics, or one may want to use a particularly good algorithm, even though it is not really fast.
A useful feature of \iffr, which can come in handy in these situations, is that it is easy to adapt the framework to parallel computation.
In particular, it is possible to distribute the calls for $\aop$, in any step, for up to $|S|$ different cores.

It is interesting to notice how the number of calls to $\aop$ relates to the input parameters 
and to the number of cycles ($n$) which the main loop executes. 
This relationship per step of the framework is shown by the following inequalities:
\begin{align*}
\mbox{$\aop$ calls on first step} & \leq n \cdot |S| \cdot (1 + 2 \cdot |E|) \\
\mbox{$\aop$ calls on second step}  & \leq n \cdot |S| \cdot p \cdot g \\
\mbox{$\aop$ calls on tail step}  & \leq |S| \cdot p \cdot t \cdot g
\end{align*}
Moreover, $n$ is closely related to other input parameters, like $i$, $b$ and $l$.

We use the instance K100 with 1000 scenarios to illustrate, in Figure~\ref{fig:improvements}, how better solutions are obtained through the cycles and steps.
In the beginning, both the first and second steps find significantly better solutions. 
After some cycles, the improvement obtained by the first step tends to zero, which we interpret as all edges worth acquiring in the first stage have already been chosen.
Meanwhile, the improvements from the second step are still significant, albeit smaller than those from the first cycles.
Eventually, the overall cycle improvement stops being significant, and then the tail step refines the best solution found, in pursuit of some final gain.

\begin{figure}[htb!]
\centering
\input{grafico_melhoria_instancia}
\caption{Improvement of best solution over time for instance K100 with 1000 scenarios.}
\label{fig:improvements}
\end{figure}

\section{Final remarks} \label{sec:conclusion}

This work provides an evolutionary framework (\iffr) that aims at solving hard two-stage stochastic resource allocation problems emerging from real-life uncertainties. 
The \iffr\ uses stage decomposition and supports the exchange of data between stages: the first stage transfers resources that were acquired in advance, while the second stage indicates resources that are frequently present among the scenarios best solutions. 

The \iffr\ was case studied using the two-stage stochastic Steiner tree problem with two sets of instances. The first set contains 24 small instances, up to 50 scenarios, previously solved by an exact algorithm \citep{Bomze2010}. The results have shown that the \iffr\ achieved near-optimal solutions, with an average gap of $1.17\%$, in a very short execution time.
The second set complies 560 instances with up to 1000 scenarios \citep{Dimacs2014}. For these instances, the \iffr\ achieved an average cost reduction of $3.8\%$ from reference solutions for which no resources are acquired in the first stage. The results attest the effectiveness of the proposed methodology in obtaining good solutions for large instances within suitable execution time.

Many sectors of modern society, such as companies, governments and academia, demand efficient solution methodologies to tackle the growing size and complexity of stochastic resource allocation problems. We believe that the \iffr\ attends this demand by providing an efficient methodology capable of solving instances that are comparable, in size and complexity, with real-life decision-making problems.

There are numerous two-stage stochastic resource allocation problems for which the \iffr\ can be applied, for example, the two-stage stochastic set cover problem. The application of the \iffr\ can be accomplished by simply implementing a heuristic for the underlying (non-stochastic) problem. Future research could investigate special-purpose improvements (regarding the problem) or even general improvements to the framework (e.g., alternative local searches).
\footnote{The complete code is available at \url{http://www.ic.unicamp.br/\textasciitilde fusberti/problems/sstp/} and does not require any commercial ILP solver.}



\bibliography{refs}

\end{document}

%% file: flow_diagram_EvFW.tex
\begin{figure}[htb!]
    \centering
    \includegraphics[width=1.0\textwidth]{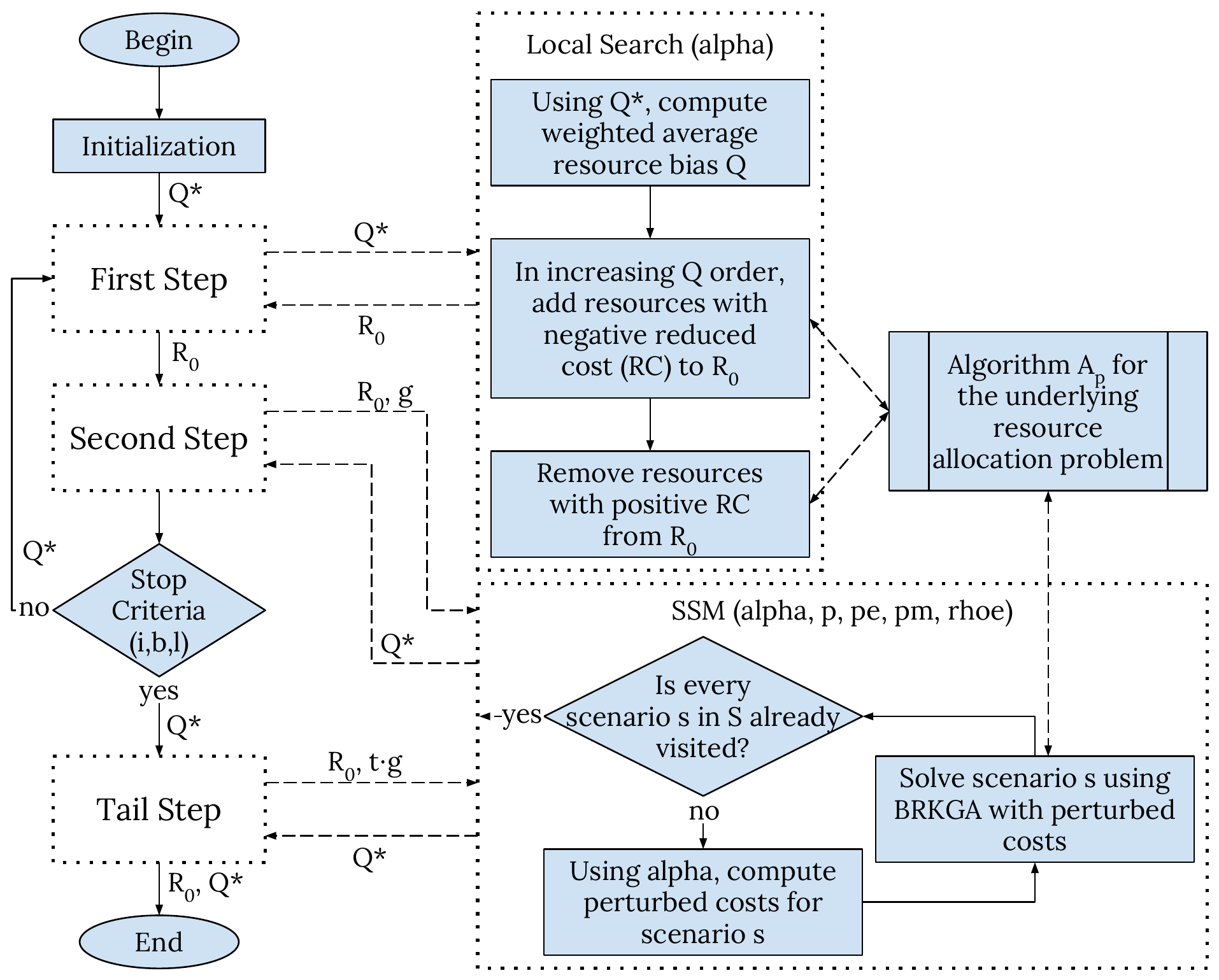}
    \caption{Flow Diagram for the \iffr.} \label{fig:Flow_Diagram_EvFW}
\end{figure}

%% file: SSTP_fig_1.tex
\begin{figure}[htb!]
\begin{center}
 
 \SetVertexSimple[
    Shape = circle,
    LineWidth = 2pt
 ]
 \SetVertexNoLabel
   \SetUpEdge[lw = 1.5pt,
    labelcolor = white,
    labelstyle = {sloped}
 ]
 
\begin{subfigure}[b]{0.4\linewidth}
\centering
\begin{tikzpicture}[scale=0.45, transform shape]
\GraphInit[vstyle=Welsh]
\LARGE
\Vertex[x=30/5 , y=40/5]{1}
\Vertex[x=37/5 , y=52/5]{2}
\Vertex[x=49/5 , y=49/5]{3}
\Vertex[x=52/5 , y=64/5]{4}
\Vertex[x=20/5 , y=26/5]{5}
\Vertex[x=40/5 , y=30/5]{6}
\Vertex[x=21/5 , y=47/5]{7}
\Vertex[x=17/5 , y=63/5]{8}
\Vertex[x=31/5 , y=62/5]{9}
\Vertex[x=52/5 , y=33/5]{10}


\Edge[label = $12$](1)(2)

\Edge[label = $13$](1)(6)
\Edge[label = $8$](1)(9)

\Edge[label = $14$](2)(4)
\Edge[label = $17$](3)(4)

\Edge[label = $16$](4)(9)
\Edge[label = $14$](5)(6)
\Edge[label = $11$](5)(7)
\Edge[label = $16$](6)(10)
\Edge[label = $9$](7)(8)
\Edge[label = $9$](8)(9)
\Edge[label = $9$](1)(7)
\Edge[label = $11$](1)(5)
\Edge[label = $20$](1)(10)
\Edge[label = $9$](3)(10)
\tikzset{EdgeStyle/.append style = {dashed}}
\end{tikzpicture}
\caption{\label{fig:1_p0}First stage}
\end{subfigure}
\begin{subfigure}[b]{0.4\linewidth}
\centering
\begin{tikzpicture}[scale=0.45, transform shape]
\GraphInit[vstyle=Welsh]
\LARGE
\Vertex[x=30/5 , y=40/5]{1}
\Vertex[x=37/5 , y=52/5]{2}
\Vertex[x=49/5 , y=49/5]{3}
\Vertex[x=52/5 , y=64/5]{4}
\Vertex[x=20/5 , y=26/5]{5}
\Vertex[x=40/5 , y=30/5]{6}
\Vertex[x=21/5 , y=47/5]{7}
\Vertex[x=17/5 , y=63/5]{8}
\Vertex[x=31/5 , y=62/5]{9}
\Vertex[x=52/5 , y=33/5]{10}

\AddVertexColor{black}{1,3, 5, 7}

\Edge[label = $13$](1)(2)
\Edge[label = $15$](1)(5)
\Edge[label = $13$](1)(6)
\Edge[label = $11$](1)(7)
\Edge[label = $9$](1)(9)
\Edge[label = $20$](1)(10)
\Edge[label = $15$](2)(4)
\Edge[label = $18$](3)(4)
\Edge[label = $13$](3)(10)
\Edge[label = $19$](4)(9)
\Edge[label = $14$](5)(6)
\Edge[label = $15$](5)(7)
\Edge[label = $18$](6)(10)
\Edge[label = $9$](7)(8)
\Edge[label = $13$](8)(9)
\tikzset{EdgeStyle/.append style = {dashed}}
\end{tikzpicture}
\caption{\label{fig:1_p1}$p_1=0.19050$}
\end{subfigure}
\begin{subfigure}[b]{0.4\linewidth}
\centering
\begin{tikzpicture}[scale=0.45, transform shape]
\GraphInit[vstyle=Welsh]
\LARGE
\Vertex[x=30/5 , y=40/5]{1}
\Vertex[x=37/5 , y=52/5]{2}
\Vertex[x=49/5 , y=49/5]{3}
\Vertex[x=52/5 , y=64/5]{4}
\Vertex[x=20/5 , y=26/5]{5}
\Vertex[x=40/5 , y=30/5]{6}
\Vertex[x=21/5 , y=47/5]{7}
\Vertex[x=17/5 , y=63/5]{8}
\Vertex[x=31/5 , y=62/5]{9}
\Vertex[x=52/5 , y=33/5]{10}

\AddVertexColor{black}{1,3, 6, 9}

\Edge[label = $12$](1)(2)
\Edge[label = $18$](1)(5)
\Edge[label = $15$](1)(6)
\Edge[label = $10$](1)(7)
\Edge[label = $8$](1)(9)
\Edge[label = $22$](1)(10)
\Edge[label = $14$](2)(4)
\Edge[label = $19$](3)(4)
\Edge[label = $14$](3)(10)
\Edge[label = $17$](4)(9)
\Edge[label = $16$](5)(6)
\Edge[label = $13$](5)(7)
\Edge[label = $17$](6)(10)
\Edge[label = $10$](7)(8)
\Edge[label = $11$](8)(9)
\tikzset{EdgeStyle/.append style = {dashed}}
\end{tikzpicture}
\caption{\label{fig:1_p2}$p_2=0.14290$}
\end{subfigure}
\begin{subfigure}[b]{0.4\linewidth}
\centering
\begin{tikzpicture}[scale=0.45, transform shape]
\GraphInit[vstyle=Welsh]
\LARGE
\Vertex[x=30/5 , y=40/5]{1}
\Vertex[x=37/5 , y=52/5]{2}
\Vertex[x=49/5 , y=49/5]{3}
\Vertex[x=52/5 , y=64/5]{4}
\Vertex[x=20/5 , y=26/5]{5}
\Vertex[x=40/5 , y=30/5]{6}
\Vertex[x=21/5 , y=47/5]{7}
\Vertex[x=17/5 , y=63/5]{8}
\Vertex[x=31/5 , y=62/5]{9}
\Vertex[x=52/5 , y=33/5]{10}

\AddVertexColor{black}{1,3, 5, 8}

\Edge[label = $14$](1)(2)
\Edge[label = $19$](1)(5)
\Edge[label = $13$](1)(6)
\Edge[label = $10$](1)(7)
\Edge[label = $9$](1)(9)
\Edge[label = $23$](1)(10)
\Edge[label = $14$](2)(4)
\Edge[label = $20$](3)(4)
\Edge[label = $13$](3)(10)
\Edge[label = $19$](4)(9)
\Edge[label = $17$](5)(6)
\Edge[label = $15$](5)(7)
\Edge[label = $18$](6)(10)
\Edge[label = $11$](7)(8)
\Edge[label = $13$](8)(9)
\tikzset{EdgeStyle/.append style = {dashed}}
\end{tikzpicture}
\caption{\label{fig:1_p3}$p_3=0.14290$}
\end{subfigure}
\begin{subfigure}[b]{0.4\linewidth}
\centering
\begin{tikzpicture}[scale=0.45, transform shape]
  \LARGE
\GraphInit[vstyle=Welsh]
\Vertex[x=30/5 , y=40/5]{1}
\Vertex[x=37/5 , y=52/5]{2}
\Vertex[x=49/5 , y=49/5]{3}
\Vertex[x=52/5 , y=64/5]{4}
\Vertex[x=20/5 , y=26/5]{5}
\Vertex[x=40/5 , y=30/5]{6}
\Vertex[x=21/5 , y=47/5]{7}
\Vertex[x=17/5 , y=63/5]{8}
\Vertex[x=31/5 , y=62/5]{9}
\Vertex[x=52/5 , y=33/5]{10}

\AddVertexColor{black}{1,5}

\Edge[label = $13$](1)(2)
\Edge[label = $16$](1)(5)
\Edge[label = $15$](1)(6)
\Edge[label = $9$](1)(7)
\Edge[label = $9$](1)(9)
\Edge[label = $23$](1)(10)
\Edge[label = $16$](2)(4)
\Edge[label = $17$](3)(4)
\Edge[label = $16$](3)(10)
\Edge[label = $17$](4)(9)
\Edge[label = $15$](5)(6)
\Edge[label = $13$](5)(7)
\Edge[label = $19$](6)(10)
\Edge[label = $10$](7)(8)
\Edge[label = $14$](8)(9)
\tikzset{EdgeStyle/.append style = {dashed}}
\end{tikzpicture}
\caption{\label{fig:1_p4}$p_4=0.30950$}
\end{subfigure}
\begin{subfigure}[b]{0.4\linewidth}
\centering
\begin{tikzpicture}[scale=0.45, transform shape]
\GraphInit[vstyle=Welsh]
\LARGE
\Vertex[x=30/5 , y=40/5]{1}
\Vertex[x=37/5 , y=52/5]{2}
\Vertex[x=49/5 , y=49/5]{3}
\Vertex[x=52/5 , y=64/5]{4}
\Vertex[x=20/5 , y=26/5]{5}
\Vertex[x=40/5 , y=30/5]{6}
\Vertex[x=21/5 , y=47/5]{7}
\Vertex[x=17/5 , y=63/5]{8}
\Vertex[x=31/5 , y=62/5]{9}
\Vertex[x=52/5 , y=33/5]{10}

\AddVertexColor{black}{1,3, 5, 8, 10}

\Edge[label = $15$](1)(2)
\Edge[label = $18$](1)(5)
\Edge[label = $14$](1)(6)
\Edge[label = $11$](1)(7)
\Edge[label = $8$](1)(9)
\Edge[label = $22$](1)(10)
\Edge[label = $16$](2)(4)
\Edge[label = $17$](3)(4)
\Edge[label = $15$](3)(10)
\Edge[label = $20$](4)(9)
\Edge[label = $16$](5)(6)
\Edge[label = $13$](5)(7)
\Edge[label = $16$](6)(10)
\Edge[label = $11$](7)(8)
\Edge[label = $13$](8)(9)
\tikzset{EdgeStyle/.append style = {dashed}}
\end{tikzpicture}
\caption{\label{fig:1_p5}$p_5=0.21420$}
\end{subfigure}

\end{center}
\caption{SSTP instance with five scenarios (terminals in black). \cite{Hokama2014}.}
\label{fig:1}
\end{figure}

%% file: SSTP_fig_2.tex
\begin{figure}[htb!]
\begin{center}
 
 \SetVertexSimple[
    Shape = circle,
    LineWidth = 2pt
 ]
 \SetVertexNoLabel
   \SetUpEdge[lw = 1.5pt,
    labelcolor = white,
    labelstyle = {sloped}
 ]
 
\begin{subfigure}[b]{0.4\linewidth}
\centering
\begin{tikzpicture}[scale=0.45, transform shape]
\GraphInit[vstyle=Welsh]
\LARGE
\Vertex[x=30/5 , y=40/5]{1}
\Vertex[x=37/5 , y=52/5]{2}
\Vertex[x=49/5 , y=49/5]{3}
\Vertex[x=52/5 , y=64/5]{4}
\Vertex[x=20/5 , y=26/5]{5}
\Vertex[x=40/5 , y=30/5]{6}
\Vertex[x=21/5 , y=47/5]{7}
\Vertex[x=17/5 , y=63/5]{8}
\Vertex[x=31/5 , y=62/5]{9}
\Vertex[x=52/5 , y=33/5]{10}


\Edge[label = $12$](1)(2)

\Edge[label = $13$](1)(6)
\Edge[label = $9$](1)(7)
\Edge[label = $8$](1)(9)
\Edge[label = $20$](1)(10)
\Edge[label = $14$](2)(4)
\Edge[label = $17$](3)(4)

\Edge[label = $16$](4)(9)
\Edge[label = $14$](5)(6)
\Edge[label = $11$](5)(7)
\Edge[label = $16$](6)(10)
\Edge[label = $9$](7)(8)
\Edge[label = $9$](8)(9)
\tikzset{EdgeStyle/.style = {line width = 2.5,blue}}
\Edge[label = $11$](1)(5)
\Edge[label = $9$](3)(10)
\tikzset{EdgeStyle/.append style = {dashed}}
\end{tikzpicture}
\caption{\label{fig:2_p0}First stage}
\end{subfigure}
\begin{subfigure}[b]{0.4\linewidth}
\centering
\begin{tikzpicture}[scale=0.45, transform shape]
\GraphInit[vstyle=Welsh]
\LARGE
\Vertex[x=30/5 , y=40/5]{1}
\Vertex[x=37/5 , y=52/5]{2}
\Vertex[x=49/5 , y=49/5]{3}
\Vertex[x=52/5 , y=64/5]{4}
\Vertex[x=20/5 , y=26/5]{5}
\Vertex[x=40/5 , y=30/5]{6}
\Vertex[x=21/5 , y=47/5]{7}
\Vertex[x=17/5 , y=63/5]{8}
\Vertex[x=31/5 , y=62/5]{9}
\Vertex[x=52/5 , y=33/5]{10}

\AddVertexColor{black}{1,3, 5, 7}

\Edge[label = $13$](1)(2)
\Edge[label = $15$](1)(5)
\Edge[label = $13$](1)(6)
\Edge[label = $11$](1)(7)
\Edge[label = $9$](1)(9)
\Edge[label = $20$](1)(10)
\Edge[label = $15$](2)(4)
\Edge[label = $18$](3)(4)
\Edge[label = $13$](3)(10)
\Edge[label = $19$](4)(9)
\Edge[label = $14$](5)(6)
\Edge[label = $15$](5)(7)
\Edge[label = $18$](6)(10)
\Edge[label = $9$](7)(8)
\Edge[label = $13$](8)(9)
\tikzset{EdgeStyle/.style = {line width = 2.5,blue}}
\Edge[label = $0$](1)(5)
\Edge[label = $0$](3)(10)
\tikzset{EdgeStyle/.style = {line width = 2.5,red}}
\Edge[label = $11$](1)(7)
\Edge[label = $20$](1)(10)
\tikzset{EdgeStyle/.append style = {dashed}}
\end{tikzpicture}
\caption{\label{fig:2_p1}$p_1=0.19050$}
\end{subfigure}
\begin{subfigure}[b]{0.4\linewidth}
\centering
\begin{tikzpicture}[scale=0.45, transform shape]
\GraphInit[vstyle=Welsh]
\LARGE
\Vertex[x=30/5 , y=40/5]{1}
\Vertex[x=37/5 , y=52/5]{2}
\Vertex[x=49/5 , y=49/5]{3}
\Vertex[x=52/5 , y=64/5]{4}
\Vertex[x=20/5 , y=26/5]{5}
\Vertex[x=40/5 , y=30/5]{6}
\Vertex[x=21/5 , y=47/5]{7}
\Vertex[x=17/5 , y=63/5]{8}
\Vertex[x=31/5 , y=62/5]{9}
\Vertex[x=52/5 , y=33/5]{10}

\AddVertexColor{black}{1,3, 6, 9}

\Edge[label = $12$](1)(2)
\Edge[label = $18$](1)(5)
\Edge[label = $15$](1)(6)
\Edge[label = $10$](1)(7)
\Edge[label = $8$](1)(9)
\Edge[label = $22$](1)(10)
\Edge[label = $14$](2)(4)
\Edge[label = $19$](3)(4)
\Edge[label = $14$](3)(10)
\Edge[label = $17$](4)(9)
\Edge[label = $16$](5)(6)
\Edge[label = $13$](5)(7)
\Edge[label = $17$](6)(10)
\Edge[label = $10$](7)(8)
\Edge[label = $11$](8)(9)
\tikzset{EdgeStyle/.style = {line width = 2.5,blue}}
\Edge[label = $0$](1)(5)
\Edge[label = $0$](3)(10)
\tikzset{EdgeStyle/.style = {line width = 2.5,red}}
\Edge[label = $8$](1)(9)
\Edge[label = $15$](1)(6)
\Edge[label = $17$](6)(10)
\tikzset{EdgeStyle/.append style = {dashed}}
\end{tikzpicture}
\caption{\label{fig:2_p2}$p_2=0.14290$}
\end{subfigure}
\begin{subfigure}[b]{0.4\linewidth}
\centering
\begin{tikzpicture}[scale=0.45, transform shape]
\GraphInit[vstyle=Welsh]
\LARGE
\Vertex[x=30/5 , y=40/5]{1}
\Vertex[x=37/5 , y=52/5]{2}
\Vertex[x=49/5 , y=49/5]{3}
\Vertex[x=52/5 , y=64/5]{4}
\Vertex[x=20/5 , y=26/5]{5}
\Vertex[x=40/5 , y=30/5]{6}
\Vertex[x=21/5 , y=47/5]{7}
\Vertex[x=17/5 , y=63/5]{8}
\Vertex[x=31/5 , y=62/5]{9}
\Vertex[x=52/5 , y=33/5]{10}

\AddVertexColor{black}{1,3, 5, 8}

\Edge[label = $14$](1)(2)
\Edge[label = $19$](1)(5)
\Edge[label = $13$](1)(6)
\Edge[label = $10$](1)(7)
\Edge[label = $9$](1)(9)
\Edge[label = $23$](1)(10)
\Edge[label = $14$](2)(4)
\Edge[label = $20$](3)(4)
\Edge[label = $13$](3)(10)
\Edge[label = $19$](4)(9)
\Edge[label = $17$](5)(6)
\Edge[label = $15$](5)(7)
\Edge[label = $18$](6)(10)
\Edge[label = $11$](7)(8)
\Edge[label = $13$](8)(9)
\tikzset{EdgeStyle/.style = {line width = 2.5,blue}}
\Edge[label = $0$](1)(5)
\Edge[label = $0$](3)(10)
\tikzset{EdgeStyle/.style = {line width = 2.5,red}}
\Edge[label = $10$](1)(7)
\Edge[label = $11$](7)(8)
\Edge[label = $23$](1)(10)
\tikzset{EdgeStyle/.append style = {dashed}}
\end{tikzpicture}
\caption{\label{fig:2_p3}$p_3=0.14290$}
\end{subfigure}
\begin{subfigure}[b]{0.4\linewidth}
\centering
\begin{tikzpicture}[scale=0.45, transform shape]
  \LARGE
\GraphInit[vstyle=Welsh]
\Vertex[x=30/5 , y=40/5]{1}
\Vertex[x=37/5 , y=52/5]{2}
\Vertex[x=49/5 , y=49/5]{3}
\Vertex[x=52/5 , y=64/5]{4}
\Vertex[x=20/5 , y=26/5]{5}
\Vertex[x=40/5 , y=30/5]{6}
\Vertex[x=21/5 , y=47/5]{7}
\Vertex[x=17/5 , y=63/5]{8}
\Vertex[x=31/5 , y=62/5]{9}
\Vertex[x=52/5 , y=33/5]{10}

\AddVertexColor{black}{1,5}

\Edge[label = $13$](1)(2)
\Edge[label = $16$](1)(5)
\Edge[label = $15$](1)(6)
\Edge[label = $9$](1)(7)
\Edge[label = $9$](1)(9)
\Edge[label = $23$](1)(10)
\Edge[label = $16$](2)(4)
\Edge[label = $17$](3)(4)
\Edge[label = $16$](3)(10)
\Edge[label = $17$](4)(9)
\Edge[label = $15$](5)(6)
\Edge[label = $13$](5)(7)
\Edge[label = $19$](6)(10)
\Edge[label = $10$](7)(8)
\Edge[label = $14$](8)(9)
\tikzset{EdgeStyle/.style = {line width = 2.5,blue}}
\Edge[label = $0$](1)(5)
\Edge[label = $0$](3)(10)
\tikzset{EdgeStyle/.append style = {dashed}}
\end{tikzpicture}
\caption{\label{fig:2_p4}$p_4=0.30950$}
\end{subfigure}
\begin{subfigure}[b]{0.4\linewidth}
\centering
\begin{tikzpicture}[scale=0.45, transform shape]
  \LARGE
\GraphInit[vstyle=Welsh]
\Vertex[x=30/5 , y=40/5]{1}
\Vertex[x=37/5 , y=52/5]{2}
\Vertex[x=49/5 , y=49/5]{3}
\Vertex[x=52/5 , y=64/5]{4}
\Vertex[x=20/5 , y=26/5]{5}
\Vertex[x=40/5 , y=30/5]{6}
\Vertex[x=21/5 , y=47/5]{7}
\Vertex[x=17/5 , y=63/5]{8}
\Vertex[x=31/5 , y=62/5]{9}
\Vertex[x=52/5 , y=33/5]{10}

\AddVertexColor{black}{1,3, 5, 8, 10}

\Edge[label = $15$](1)(2)
\Edge[label = $18$](1)(5)
\Edge[label = $14$](1)(6)
\Edge[label = $11$](1)(7)
\Edge[label = $8$](1)(9)
\Edge[label = $22$](1)(10)
\Edge[label = $16$](2)(4)
\Edge[label = $17$](3)(4)
\Edge[label = $15$](3)(10)
\Edge[label = $20$](4)(9)
\Edge[label = $16$](5)(6)
\Edge[label = $13$](5)(7)
\Edge[label = $16$](6)(10)
\Edge[label = $11$](7)(8)
\Edge[label = $13$](8)(9)
\tikzset{EdgeStyle/.style = {line width = 2.5,blue}}
\Edge[label = $0$](1)(5)
\Edge[label = $0$](3)(10)
\tikzset{EdgeStyle/.style = {line width = 2.5,red}}
\Edge[label = $8$](1)(9)
\Edge[label = $13$](8)(9)
\Edge[label = $22$](1)(10)
\tikzset{EdgeStyle/.append style = {dashed}}
\end{tikzpicture}
\caption{\label{fig:2_p5}$p_5=0.21420$}
\end{subfigure}

\end{center}
\caption{Solution of the first (blue) and second (red) stages. \cite{Hokama2014}.}
\label{fig:2}
\end{figure}

%% file: BRKGA_fig_1.tex
\begin{figure}[htb!]
\begin{center}

 \SetVertexSimple[
  Shape = circle,
  LineWidth = 2pt
 ]
 \SetVertexNoLabel
 \SetUpEdge[
  lw = 1.5pt,
  labelcolor = white,
  labelstyle = {sloped}
 ]

\begin{subfigure}[b]{0.4\linewidth}
\centering
\begin{tikzpicture}[scale=0.45, transform shape]
\GraphInit[vstyle=Welsh]
\LARGE
\Vertex[x=15/5 , y=15/5]{1}
\Vertex[x=45/5 , y=15/5]{2}
\Vertex[x=30/5 , y=27/5]{3}
\Vertex[x=30/5 , y=45/5]{4}

\AddVertexColor{black}{1,2,4}

\Edge[label = $f$](1)(2)
\Edge[label = $d$](1)(3)
\Edge[label = $a$](1)(4)
\Edge[label = $e$](2)(3)
\Edge[label = $c$](2)(4)
\Edge[label = $b$](3)(4)

\tikzset{EdgeStyle/.style = {line width = 2.5,blue}}
\tikzset{EdgeStyle/.append style = {dashed}}
\end{tikzpicture}
\caption{\label{fig:decoder_a}}
\end{subfigure}
\begin{subfigure}[b]{0.4\linewidth}
\centering
\begin{tikzpicture}[scale=0.45, transform shape]
\GraphInit[vstyle=Welsh]
\LARGE
\Vertex[x=15/5 , y=15/5]{1}
\Vertex[x=45/5 , y=15/5]{2}
\Vertex[x=30/5 , y=27/5]{3}
\Vertex[x=30/5 , y=45/5]{4}

\AddVertexColor{black}{1,2, 4}

\Edge[label = $5$](1)(3)
\Edge[label = $5$](2)(3)
\Edge[label = $9$](2)(4)
\Edge[label = $5$](3)(4)

\tikzset{EdgeStyle/.style = {line width = 2.5,blue}}
\Edge[label = $9$](1)(2)
\Edge[label = $9$](1)(4)
\tikzset{EdgeStyle/.append style = {dashed}}
\end{tikzpicture}
\caption{\label{fig:decoder_b}}
\end{subfigure}
\begin{subfigure}[b]{0.4\linewidth}
\centering
\begin{tikzpicture}[scale=0.45, transform shape]]
\GraphInit[vstyle=Welsh]
\LARGE
\Vertex[x=15/5 , y=15/5]{1}
\Vertex[x=45/5 , y=15/5]{2}
\Vertex[x=30/5 , y=27/5]{3}
\Vertex[x=30/5 , y=45/5]{4}

\AddVertexColor{black}{1,2, 4}

\Edge[label = $9$](2)(4)
\Edge[label = $9.9$](1)(2)
\Edge[label = $10.8$](1)(4)
\tikzset{EdgeStyle/.style = {line width = 2.5,blue}}
\Edge[label = $5$](1)(3)
\Edge[label = $4.5$](3)(4)
\Edge[label = $4$](2)(3)
\tikzset{EdgeStyle/.append style = {dashed}}
\end{tikzpicture}
\caption{\label{fig:decoder_c}}
\end{subfigure}

\end{center}
\caption{Solution improved by \brkga\ perturbed costs. \cite{Hokama2014}.}
\label{fig:decoder}
\end{figure}

%% file: graficos_alpha_rhoe.tex
\begin{figure}[htb!]
  \centering
  \begin{subfigure}{.5\textwidth}
    \centering
    \begin{tikzpicture}
			\tiny
			\selectcolormodel{gray}
			\begin{axis}[xlabel=$\alpha$, ylabel=Improvement(\%),
			      enlargelimits=0.15, extra x ticks={0.5,0.6,0.7,0.8,0.9},
			      height=4cm,width=\textwidth, grid=major, /pgf/number format/precision=3]
			  \addplot coordinates {
			    (0.5,	3.8947)
			    (0.6,	3.8978)
			    (0.7,	3.9010)
			    (0.8,	3.8911)
			    (0.9,	3.8845)
			  };
			\end{axis}
	  \end{tikzpicture}
    \caption{$\alpha$ convergence.} \label{fig:alpha}
  \end{subfigure}%
  \begin{subfigure}{.5\textwidth}
		\begin{tikzpicture}
			\tiny
			\selectcolormodel{gray}
			\begin{axis}[xlabel=$\rho_e$, ylabel=,
			      enlargelimits=0.15, extra x ticks={0.2,0.3,0.4,0.5,0.6},
			      height=4cm,width=\textwidth, grid=major]
			  \addplot coordinates {
			    (0.2,	3.8704)
			    (0.3,	3.8859)
			    (0.4,	3.9010)
			    (0.5,	3.8920)
			    (0.6,	3.8817)
			  };
			\end{axis}
		\end{tikzpicture}
        \caption{$\rho_e$ convergence.} \label{fig:rhoe}
    \end{subfigure}
    \caption{Parameters local convergence.} \label{fig:param_local_convergence}
\end{figure}
%
%

%% file: graficos_pop_blt.tex
\begin{figure}[htb!]
  \centering
  \begin{subfigure}{.5\textwidth}
    \centering
    \begin{tikzpicture}
			\tiny
			\selectcolormodel{gray}
			\begin{axis}[xlabel=Time ratio, ylabel=Improvement(\%),
			      enlargelimits=0.15, extra x ticks={0.5,1,1.5,2,2.5,3,3.5},
			      height=4cm,width=\textwidth, grid=major]
			  \addplot coordinates {
					(0.6497, 3.8517)
					(1.0000, 3.9010)
					(1.6787, 3.9304)
					(3.0245, 3.9475)
			  };
		  \node[label={0:{\tiny p = 12}},inner sep=1pt, ] at (axis cs:0.6497,3.8517) {};
		  \node[label={320:{\tiny 25}},inner sep=1pt, ] at (axis cs:1.0000, 3.9010) {};
		  \node[label={90:{\tiny 50}},inner sep=1pt, ] at (axis cs:1.6787,3.9304) {};
		  \node[label={270:{\tiny 100}},inner sep=1pt, ] at (axis cs:3.0245,3.9475) {};
			\end{axis}
	  \end{tikzpicture}
    \caption{Population convergence.} \label{fig:pop}
  \end{subfigure}%
  \begin{subfigure}{.5\textwidth}
		\begin{tikzpicture}
			\tiny
			\selectcolormodel{gray}
			\begin{axis}[xlabel=Time ratio, ylabel=,
			      enlargelimits=0.15, extra x ticks={0.5,1,1.5,2,2.5,3,3.5},
			      height=4cm,width=\textwidth, grid=major]
			  \addplot coordinates {
					(0.1814, 3.5060)
					(0.5410, 3.8196)
					(1.0000, 3.9010)
					(1.7901, 3.9586)
					(2.7160, 3.9861)
			  };
		  \node[label={0:{\tiny b/l/t = 0/0/0}},inner sep=1pt, ] at (axis cs:0.1814,3.5060) {};
		  \node[label={320:{\tiny 1/1/1}},inner sep=1pt, ] at (axis cs:0.5410,3.8196) {};
		  \node[label={90:{\tiny 3/2/3}},inner sep=1pt, ] at (axis cs:1.0000,3.9010) {};
		  \node[label={270:{\tiny 6/4/6}},inner sep=1pt, ] at (axis cs:1.7901,3.9586) {};
		  \node[label={270:{\tiny 10/6/10}},inner sep=1pt, ] at (axis cs:2.7160,3.9861) {};
			\end{axis}
		\end{tikzpicture}
        \caption{$b$, $l$, $t$ convergence.} \label{fig:blt}
    \end{subfigure}
    \caption{Parameters asymptotic convergence.} \label{fig:param_asymptotic_convergence}
\end{figure}
%
%

%% file: tabela_bomze.tex
\begin{table}[htb!]
\renewcommand\tabcolsep{5pt}
\begin{center}
\caption{Results for instances from benchmark 1.} \label{tab:Bomze}
\begin{tabular}{l r r | r r | r r| r r r}
 &  &  &  \multicolumn{2}{c}{$EF$} & \multicolumn{2}{|c}{$2BC^*$} & \multicolumn{2}{|c}{\iffr} \\ 
\multicolumn{1}{c}{instance} & \multicolumn{1}{c}{$|S|$} 	& \multicolumn{1}{c}{$OPT$} &  \multicolumn{1}{|c}{time} 	 & 	\multicolumn{1}{c}{gap}	 & \multicolumn{1}{|c}{time} 	 & 	\multicolumn{1}{c}{gap}	  & 	 \multicolumn{1}{|c}{time} 	 & 	\multicolumn{1}{c}{gap}  	\\ \hline
lin01\_53\_80 & 5 & 797.0 & 0.2 & - & 2.2 & - & 8.4 & 2.9 \\
lin01\_53\_80 & 10 & 633.2 & 0.7 & - & 2.5 & - & 10.9 & 0.9 \\
lin01\_53\_80 & 20 & 753.9 & 5.7 & - & 6.9 & - & 29.4 & 0.3 \\
lin01\_53\_80 & 50 & 768.9 & 33.4 & - & 10.4 & - & 75.6 & 0.1 \\
lin02\_55\_82 & 5 & 476.2 & 0.1 & - & 1.1 & - & 4.6 & 0.0 \\
lin02\_55\_82 & 10 & 739.1 & 1.0 & - & 3.0 & - & 16.5 & 1.7 \\
lin02\_55\_82 & 20 & 752.2 & 4.9 & - & 4.3 & - & 31.1 & 0.3 \\
lin02\_55\_82 & 50 & 732.6 & 31.2 & - & 10.7 & - & 76.9 & 0.5 \\
lin03\_57\_84 & 5 & 653.0 & 0.5 & - & 1.9 & - & 6.0 & 0.7 \\
lin03\_57\_84 & 10 & 834.7 & 3.8 & - & 8.7 & - & 19.2 & 1.4 \\
lin03\_57\_84 & 20 & 854.9 & 10.8 & - & 7.3 & - & 38.7 & 0.1 \\
lin03\_57\_84 & 50 & 895.7 & 103.1 & - & 21.3 & - & 96.7 & 0.2 \\
lin04\_157\_266 & 5 & 1922.1 & 140.4 & - & 959.2 & - & 84.6 & 1.6 \\
lin04\_157\_266 & 10 & 1959.1 & 415.8 & - & 989.2 & - & 139.9 & 0.6 \\
lin04\_157\_266 & 20 & 1954.9 & 5498.7 & - & 3016.7 & - & 239.2 & 1.4 \\
lin04\_157\_266 & 50 & 2097.7 & 7200.0 & 19.5 & 5330.2 & - & 546.3 & 4.3 \\
lin05\_160\_269 & 5 & 2215.5 & 282.0 & - & 2681.2 & - & 108.3 & 2.6 \\
lin05\_160\_269 & 10 & 2210.2 & 1866.7 & - & 4096.0 & - & 174.3 & 3.1 \\
lin05\_160\_269 & 20 & {\bf 2412.2$^\diamond$} & 7200.0 & 5.6 & 7200.0 & 4.7 & 352.9 & -0.8 \\
lin05\_160\_269 & 50 & 2297.0 & 7200.0 & 21.3 & 3627.4 & - & 842.9 & 2.0 \\
lin06\_165\_274 & 5 & 1975.8 & 212.8 & - & 760.9 & - & 88.0 & 1.0 \\
lin06\_165\_274 & 10 & 1918.7 & 501.7 & - & 808.4 & - & 126.4 & 0.2 \\
lin06\_165\_274 & 20 & 2457.6 & 7200.0 & - & 3222.9 & - & 473.7 & 2.5 \\
lin06\_165\_274 & 50 & 2186.8 & 7200.0 & 22.5 & 2795.5 & - & 1139.3 & 0.5 \\
\hline
Average & & & 1879.7 & & 1482.0 & & 197.1 & \\
\hline
\multicolumn{9}{l}{$\diamond$ instance in which \cite{Bomze2010} does not reach the optimum.} \\
\multicolumn{9}{l}{In fact, we found a solution with cost 2356.5.} \\
\multicolumn{9}{l}{$|S|$ -- number of scenarios.} \\
\multicolumn{9}{l}{$OPT$ -- Optimal solution costs obtained by \cite{Bomze2010}.} \\
\multicolumn{9}{l}{$EF$ -- results from Extensive Form algorithm by \cite{Bomze2010}.} \\
\multicolumn{9}{l}{$2BC^*$ -- results from 2-stage B\&C algorithm by \cite{Bomze2010}.} \\
\multicolumn{9}{l}{\iffr -- average results from evolutionary framework.} \\
\multicolumn{9}{l}{time -- computational time in seconds.} \\
\multicolumn{9}{l}{gap -- optimality gap in percentage.} \\
\end{tabular}
\end{center}
\end{table}

%% file: tabela_DIMACS.tex
\afterpage{
\renewcommand\tabcolsep{4pt}
\begin{landscape}
\begin{table}
\begin{center}
\small
\caption{Summarized results for instances from benchmark 2.}  \label{tab:Dimacs}
\begin{tabular}{r r r r r r r r r r r r r r r r r r r r r r r r r r r}
 &  & \multicolumn{5}{c}{$K$}  &  & \multicolumn{5}{c}{$Lin$} &  & \multicolumn{5}{c}{$P$} &  & \multicolumn{5}{c}{$Wrp$} \\ \cline{3-7} \cline{9-13} \cline{15-19} \cline{21-25}
 &  & \multicolumn{2}{c}{$\delta c$ (\%)} &  & \multicolumn{2}{c}{time (s)} &  & \multicolumn{2}{c}{$\delta c$ (\%)} &  & \multicolumn{2}{c}{time (s)} &  & \multicolumn{2}{c}{$\delta c$ (\%)} &  & \multicolumn{2}{c}{time (s)} &  & \multicolumn{2}{c}{$\delta c$ (\%)} &  & \multicolumn{2}{c}{time (s)} \\ \cline{3-4} \cline{6-7} \cline{9-10} \cline{12-13} \cline{15-16} \cline{18-19} \cline{21-22} \cline{24-25}
$|S|$ &  & MH & \iffr &  & MH & \iffr &  & MH & \iffr &  & MH & \iffr &  & MH & \iffr &  & MH & \iffr &  & MH & \iffr &  & MH & \iffr \\ \hline
5 &  & 6.93 & 5.72 &  & 522.3 & 10.2 &  & 6.64 & 6.55 &  & 2167.3 & 210.7 &  & 5.50 & 4.85 &  & 1434.6 & 44.0 &  & 3.34 & 3.23 &  & 1540.9 & 191.2 \\ 
10 &  & 5.90 & 5.34 &  & 654.2 & 22.4 &  & 4.88 & 5.22 &  & 1441.5 & 416.7 &  & 4.81 & 4.59 &  & 717.5 & 89.5 &  & 2.86 & 2.79 &  & 1720.4 & 399.9 \\ 
20 &  & 5.03 & 4.72 &  & 1323.7 & 38.8 &  & 4.78 & 5.07 &  & 3274.3 & 688.7 &  & 4.55 & 4.30 &  & 2597.3 & 184.8 &  & 2.79 & 2.72 &  & 2703.5 & 731.2 \\ 
50 &  & 3.90 & 4.06 &  & 3272.7 & 91.4 &  & 4.19 & 4.80 &  & 4538.0 & 1451.9 &  & 3.62 & 3.99 &  & 2877.7 & 445.6 &  & 2.71 & 2.65 &  & 3517.1 & 1794.6 \\ 
75 &  & 4.02 & 4.09 &  & 3244.3 & 137.0 &  & 3.73 & 4.58 &  & 5390.2 & 2089.3 &  & 3.70 & 4.26 &  & 3375.6 & 644.1 &  & 2.67 & 2.60 &  & 4252.5 & 2658.7 \\ 
100 &  & 3.97 & 3.96 &  & 3818.2 & 182.7 &  & 3.32 & 4.48 &  & 4986.4 & 2777.0 &  & 3.73 & 4.37 &  & 4098.8 & 856.2 &  & 2.64 & 2.58 &  & 4526.2 & 3378.0 \\ 
150 &  & 3.77 & 3.78 &  & 4485.3 & 264.0 &  & 3.21 & 4.46 &  & 5308.5 & 4257.0 &  & 3.91 & 4.49 &  & 5393.9 & 1299.2 &  & 2.62 & 2.59 &  & 5239.6 & 5164.5 \\ 
\hline
Avg. &  & 4.79 & 4.52 &  & 2474.4 & 106.6 &  & 4.39 & 5.02 &  & 3872.3 & 1698.8 &  & 4.26 & 4.41 &  & 2927.9 & 509.1 &  & 2.80 & 2.74 &  & 3357.2 & 2045.4 \\
\hline
200 &  & 3.80 & 3.87 &  & 5270.2 & 37.5 &  & 2.99 & 4.38 &  & 5269.0 & 1008.1 &  & 3.69 & 4.35 &  & 5530.2 & 196.8 &  & 2.58 & 2.55 &  & 5345.8 & 841.2 \\ 
250 &  & 3.89 & 3.88 &  & 5088.7 & 47.7 &  & 2.97 & 4.33 &  & 5424.8 & 1243.3 &  & 3.68 & 4.33 &  & 5708.5 & 257.6 &  & 2.53 & 2.53 &  & 5524.4 & 1067.2 \\ 
300 &  & 3.73 & 3.85 &  & 5531.0 & 59.5 &  & 2.76 & 4.29 &  & 5504.3 & 1509.7 &  & 3.68 & 4.46 &  & 5585.2 & 301.0 &  & 2.50 & 2.53 &  & 5650.5 & 1263.2 \\ 
400 &  & 3.52 & 3.70 &  & 5599.1 & 77.6 &  & 2.55 & 4.36 &  & 5652.8 & 2043.1 &  & 3.61 & 4.40 &  & 5687.3 & 392.4 &  & 2.46 & 2.53 &  & 5672.4 & 1661.5 \\ 
500 &  & 3.56 & 3.89 &  & 5659.9 & 96.0 &  & 2.38 & 4.37 &  & 5666.0 & 2476.6 &  & 3.49 & 4.39 &  & 5683.2 & 517.7 &  & 2.43 & 2.54 &  & 5713.7 & 2055.3 \\ 
750 &  & 3.41 & 4.02 &  & 5640.4 & 145.6 &  & 2.16 & 4.41 &  & 5734.1 & 3706.9 &  & 3.06 & 4.32 &  & 5694.4 & 742.7 &  & 2.37 & 2.54 &  & 5606.4 & 3112.8 \\ 
1000 &  & 3.22 & 3.94 &  & 5694.7 & 191.4 &  & 1.96 & 4.38 &  & 5721.3 & 4955.4 &  & 2.99 & 4.25 &  & 5707.2 & 936.2 &  & 2.34 & 2.57 &  & 5671.5 & 4145.8 \\ 
\hline
Avg. &  & 3.59 & 3.88 &  & 5497.7 & 93.6 &  & 2.54 & 4.36 &  & 5567.5 & 2420.4 &  & 3.46 & 4.36 &  & 5656.6 & 477.8 &  & 2.46 & 2.54 &  & 5597.8 & 2021.0 \\
\hline
\multicolumn{25}{l}{$K, Lin, P, Wrp$ -- DIMACS instances for the SSTP, available at \url{ http://dimacs11.zib.de/downloads.html}\,.}\\
\multicolumn{25}{l}{$\delta c$ -- average improvement ratio relative to the \textit{buy\_none} solutions.}  \\
\multicolumn{25}{l}{time -- average execution time (in seconds) to achieve the best solution. For fairness, MH time was corrected by a $1.6$ factor.}  \\
\multicolumn{25}{l}{$|S|$ -- number of scenarios.}  \\
\multicolumn{25}{l}{MH -- heuristic results from \cite{Hokama2014}.} \\
\multicolumn{25}{l}{\iffr -- average results from evolutionary framework.} \\
\end{tabular}
\end{center}
\end{table}
\end{landscape}
}

%% file: graficos_tempo_passos.tex
\newcommand{\wheelchart}[3]{
    \pgfmathsetmacro{\totalnum}{0}
    \foreach \value/\colour in {#1} {
        \pgfmathparse{\value+\totalnum}
        \global\let\totalnum=\pgfmathresult
    }

    \pgfmathsetmacro{\wheelwidth}{(#3)-(#2)}
    \pgfmathsetmacro{\midradius}{(#3+#2)/2}

    \begin{scope}[rotate=90]
        \pgfmathsetmacro{\cumnum}{0}
        \foreach \value/\colour in {#1} {
            \pgfmathsetmacro{\newcumnum}{\cumnum + \value/\totalnum*360}

            \draw[fill=\colour, line width=1mm, draw=white] (-\cumnum:#2) arc (-\cumnum:-\newcumnum:#2)--(-\newcumnum:#3) arc (-\newcumnum:-\cumnum:#3)--cycle;

            \global\let\cumnum=\newcumnum
      }
      \end{scope}
}

\begin{figure}[H]
  \centering
  \begin{subfigure}{\textwidth}
    \centering
    \begin{tikzpicture}
      \tiny
      \draw[draw=black] (-2mm, -6.5mm) rectangle (81mm, 3.5mm);
      \filldraw[fill=blue!30, draw=black] (0,0) rectangle (6mm,2mm) node[right, black, yshift=-1mm] {First step time};
      \filldraw[fill=red!30, draw=black] (28mm,0) rectangle (34mm,2mm) node[right, black, yshift=-1mm] {Second step time};
      \filldraw[fill=yellow!30, draw=black] (57mm,0) rectangle (63mm,2mm) node[right, black, yshift=-1mm] {Tail step time};
      \filldraw[fill=green!30, draw=black] (17mm,-5mm) rectangle (23mm,-3mm) node[right, black, yshift=-1mm] {$\aop$ time};
      \filldraw[fill=magenta!30, draw=black] (47mm,-5mm) rectangle (53mm,-3mm) node[right, black, yshift=-1mm] {\iffr\ time};
    \end{tikzpicture}
  \end{subfigure}
  
  \vspace{2mm}
  
  \begin{center}
  \begin{subfigure}{.4\textwidth}
  \centering
    \begin{tikzpicture}[scale=0.7]
      \wheelchart{91.87/green!30, 8.13/magenta!30}{0cm}{1cm}
      \wheelchart{8.99/blue!30, 44.92/red!30, 46.07/yellow!30}{1.02cm}{2.02cm}
    \end{tikzpicture}
    \caption{Small instances with more permissive parameters.} \label{fig:ratio_pequenas}
  \end{subfigure}%
  \hspace{0.1\textwidth}
  \begin{subfigure}{.4\textwidth}
  \centering
    \begin{tikzpicture}[scale=0.7]
      \wheelchart{92.23/green!30, 7.77/magenta!30}{0cm}{1cm}
      \wheelchart{28.55/blue!30, 40.14/red!30, 31.23/yellow!30}{1.02cm}{2.02cm}
     \end{tikzpicture}
     \caption{Large instances with standard parameters.} \label{fig:ratio_grandes}
  \end{subfigure}%
  \end{center}
  \caption{Time ratio per step and while solving $A_p$ (MST-approx).} \label{fig:time_ratio_step}
\end{figure}


%% file: grafico_melhoria_instancia.tex
\begin{tikzpicture}
\tiny
\begin{axis}[
      xlabel=Time (minutes),
      ylabel=Improvement(\%),
      ymax=5.2,
      enlargelimits=0.05,
      height=6cm,width=\textwidth,
      clip=false,
      legend style={at={(0.5,1.1)},anchor=south},
      legend columns=-1,
      grid=major
      ]

    \addplot coordinates {
(0.000000,	0.000000)
(0.273000,	0.511975)
(0.777383,	2.753795)
(1.072743,	3.219451)
(1.587805,	4.407060)
(1.888017,	4.408295)
(2.404067,	4.601598)
(2.705367,	4.602216)
(3.222900,	4.708440)
(3.523900,	4.708440)
(4.041267,	4.766493)
(4.342450,	4.767110)
(4.860067,	4.803547)
(6.326250,	4.866541)
      }; 
      
    \def \LScolor {blue!100};
    \def \SScolor {red!100};
    \def \TScolor {yellow!100};
    \addplot [name path=B, white!0] coordinates { (0,5.1) (6.5,5.1) };  
    \addplot [name path=A, white!0] coordinates { (0,-0.1) (6.5,-0.1) };  
    
    \addplot [fill = \LScolor, fill opacity=.3, draw = none, area legend] fill between [of=A and B, soft clip={domain=0.000000:0.273000}];          
    \addplot [fill = \SScolor, fill opacity=.3, draw = none, area legend] fill between [of=A and B, soft clip={domain=0.273000:0.7773834}];  
    \addplot [fill = \TScolor, fill opacity=.3, draw = none, area legend] fill between [of=A and B, soft clip={domain=4.860067:6.326250}];
    \addplot [\LScolor, fill opacity=.3] fill between [of=A and B, soft clip={domain=0.777383:1.072743}];
    \addplot [\SScolor, fill opacity=.3] fill between [of=A and B, soft clip={domain=1.072743:1.587805}];
    \addplot [\LScolor, fill opacity=.3] fill between [of=A and B, soft clip={domain=1.587805:1.888017}];
    \addplot [\SScolor, fill opacity=.3] fill between [of=A and B, soft clip={domain=1.888017:2.404067}];
    \addplot [\LScolor, fill opacity=.3] fill between [of=A and B, soft clip={domain=2.404067:2.705367}];
    \addplot [\SScolor, fill opacity=.3] fill between [of=A and B, soft clip={domain=2.705367:3.222900}];
    \addplot [\LScolor, fill opacity=.3] fill between [of=A and B, soft clip={domain=3.222900:3.523900}];
    \addplot [\SScolor, fill opacity=.3] fill between [of=A and B, soft clip={domain=3.523900:4.041267}];
    \addplot [\LScolor, fill opacity=.3] fill between [of=A and B, soft clip={domain=4.041267:4.342450}];
    \addplot [\SScolor, fill opacity=.3] fill between [of=A and B, soft clip={domain=4.342450:4.860067}];

\legend{,,,First step, Second step, Tail step}
\end{axis}

\end{tikzpicture}

%% file: source_file.bbl
\begin{thebibliography}{34}
\expandafter\ifx\csname natexlab\endcsname\relax\def\natexlab#1{#1}\fi
\providecommand{\bibinfo}[2]{#2}
\ifx\xfnm\relax \def\xfnm[#1]{\unskip,\space#1}\fi
\bibitem[{Amorim et~al.(2015)Amorim, Costa \&
  Almada-Lobo}]{AmorimCostaAlmada2015}
\bibinfo{author}{Amorim, P.}, \bibinfo{author}{Costa, A.~M.}, \&
  \bibinfo{author}{Almada-Lobo, B.} (\bibinfo{year}{2015}).
\newblock \bibinfo{title}{A hybrid path-relinking method for solving two-stage
  stochastic integer problems}.
\newblock {\it \bibinfo{journal}{International Transactions in Operational
  Research}\/},  {\it \bibinfo{volume}{22}\/}, \bibinfo{pages}{113--127}.
\bibitem[{Badri et~al.(2017)Badri, Fatemi~Ghomi \& Hejazi}]{Badri20171}
\bibinfo{author}{Badri, H.}, \bibinfo{author}{Fatemi~Ghomi, S.}, \&
  \bibinfo{author}{Hejazi, T.-H.} (\bibinfo{year}{2017}).
\newblock \bibinfo{title}{A two-stage stochastic programming approach for
  value-based closed-loop supply chain network design}.
\newblock {\it \bibinfo{journal}{Transportation Research Part E: Logistics and
  Transportation Review}\/},  {\it \bibinfo{volume}{105}\/},
  \bibinfo{pages}{1--17}.
\bibitem[{Birge \& Louveaux(2011)}]{birge2011introduction}
\bibinfo{author}{Birge, J.~R.}, \& \bibinfo{author}{Louveaux, F.}
  (\bibinfo{year}{2011}).
\newblock {\it \bibinfo{title}{Introduction to stochastic programming}\/}.
\newblock \bibinfo{publisher}{Springer Science \& Business Media}.
\bibitem[{Bomze et~al.(2010)Bomze, Chimani, J{\"u}nger, Ljubi{\'{c}}, Mutzel \&
  Zey}]{Bomze2010}
\bibinfo{author}{Bomze, I.}, \bibinfo{author}{Chimani, M.},
  \bibinfo{author}{J{\"u}nger, M.}, \bibinfo{author}{Ljubi{\'{c}}, I.},
  \bibinfo{author}{Mutzel, P.}, \& \bibinfo{author}{Zey, B.}
  (\bibinfo{year}{2010}).
\newblock \bibinfo{title}{Solving two-stage stochastic steiner tree problems by
  two-stage branch-and-cut}.
\newblock In {\it \bibinfo{booktitle}{Algorithms and Computation: 21st
  International Symposium (ISAAC 2010), Proceedings, Part I}\/} (pp.
  \bibinfo{pages}{427--439}).
\newblock \bibinfo{publisher}{Springer Berlin Heidelberg} volume
  \bibinfo{volume}{6506} of {\it \bibinfo{series}{Lecture Notes in Computer
  Science}\/}.
\bibitem[{Caceres-Cruz et~al.(2014)Caceres-Cruz, Arias, Guimarans, Riera \&
  Juan}]{Caceres-Cruz:2014}
\bibinfo{author}{Caceres-Cruz, J.}, \bibinfo{author}{Arias, P.},
  \bibinfo{author}{Guimarans, D.}, \bibinfo{author}{Riera, D.}, \&
  \bibinfo{author}{Juan, A.~A.} (\bibinfo{year}{2014}).
\newblock \bibinfo{title}{Rich vehicle routing problem: Survey}.
\newblock {\it \bibinfo{journal}{ACM Comput. Surv.}\/},  {\it
  \bibinfo{volume}{47}\/}, \bibinfo{pages}{32:1--32:28}.
\bibitem[{Carreira et~al.(2017)Carreira, Lulli \& Antunes}]{Carreira2017639}
\bibinfo{author}{Carreira, J.}, \bibinfo{author}{Lulli, G.}, \&
  \bibinfo{author}{Antunes, A.} (\bibinfo{year}{2017}).
\newblock \bibinfo{title}{The airline long-haul fleet planning problem: The
  case of tap service to/from brazil}.
\newblock {\it \bibinfo{journal}{European Journal of Operational Research}\/},
  {\it \bibinfo{volume}{263}\/}, \bibinfo{pages}{639--651}.
\bibitem[{Cavalli-Sforza \& Edwards(1967)}]{CavalliEdwards1967}
\bibinfo{author}{Cavalli-Sforza, L.~L.}, \& \bibinfo{author}{Edwards, A.}
  (\bibinfo{year}{1967}).
\newblock \bibinfo{title}{Phylogenetic analysis. models and estimation
  procedures}.
\newblock {\it \bibinfo{journal}{American Journal of Human Genetics}\/},  {\it
  \bibinfo{volume}{19(3)}\/}, \bibinfo{pages}{233--257}.
\bibitem[{Christofides \& Korman(1975)}]{ChristofidesKorman1975}
\bibinfo{author}{Christofides, N.}, \& \bibinfo{author}{Korman, S.}
  (\bibinfo{year}{1975}).
\newblock \bibinfo{title}{Note—a computational survey of methods for the set
  covering problem}.
\newblock {\it \bibinfo{journal}{Management Science}\/},  {\it
  \bibinfo{volume}{21}\/}, \bibinfo{pages}{591--599}.
\bibitem[{Cobuloglu \& Esra~Buyuktahtakin(2017)}]{Cobuloglu2017251}
\bibinfo{author}{Cobuloglu, H.}, \& \bibinfo{author}{Esra~Buyuktahtakin, I.}
  (\bibinfo{year}{2017}).
\newblock \bibinfo{title}{A two-stage stochastic mixed-integer programming
  approach to the competition of biofuel and food production}.
\newblock {\it \bibinfo{journal}{Computers and Industrial Engineering}\/},
  {\it \bibinfo{volume}{107}\/}, \bibinfo{pages}{251--263}.
\bibitem[{Cunha et~al.(2017)Cunha, Raupp \& Oliveira}]{Cunha2017313}
\bibinfo{author}{Cunha, P.}, \bibinfo{author}{Raupp, F.}, \&
  \bibinfo{author}{Oliveira, F.} (\bibinfo{year}{2017}).
\newblock \bibinfo{title}{A two-stage stochastic programming model for periodic
  replenishment control system under demand uncertainty}.
\newblock {\it \bibinfo{journal}{Computers and Industrial Engineering}\/},
  {\it \bibinfo{volume}{107}\/}, \bibinfo{pages}{313--326}.
\bibitem[{DIMACS(2014)}]{Dimacs2014}
\bibinfo{author}{DIMACS} (\bibinfo{year}{2014}).
\newblock \bibinfo{title}{11th {DIMACS} {I}mplementation {C}hallenge in
  {C}ollaboration with {ICERM}: {S}teiner {T}ree {P}roblems}.
\bibitem[{Drezner(1995)}]{Drezner1995}
\bibinfo{author}{Drezner, Z.~E.} (\bibinfo{year}{1995}).
\newblock {\it \bibinfo{title}{Facility Location: A Survey of Applications and
  Methods}\/}.
\newblock Springer Series in Operations Research and Financial Engineering.
\newblock \bibinfo{publisher}{Springer}.
\bibitem[{Gon{\c{c}}alves \& Resende(2011)}]{Goncalves2011}
\bibinfo{author}{Gon{\c{c}}alves, J.~F.}, \& \bibinfo{author}{Resende, M.
  G.~C.} (\bibinfo{year}{2011}).
\newblock \bibinfo{title}{Biased random-key genetic algorithms for
  combinatorial optimization}.
\newblock {\it \bibinfo{journal}{Journal of Heuristics}\/},  {\it
  \bibinfo{volume}{17}\/}, \bibinfo{pages}{487--525}.
\bibitem[{Gupta et~al.(2007)Gupta, Ravi \& Sinha}]{Gupta07}
\bibinfo{author}{Gupta, A.}, \bibinfo{author}{Ravi, R.}, \&
  \bibinfo{author}{Sinha, A.} (\bibinfo{year}{2007}).
\newblock \bibinfo{title}{Lp rounding approximation algorithms for stochastic
  network design}.
\newblock {\it \bibinfo{journal}{Mathematics of Operations Research}\/},  {\it
  \bibinfo{volume}{32}\/}, \bibinfo{pages}{345--364}.
\bibitem[{Hokama et~al.(2014)Hokama, San~Felice, Bracht \&
  Usberti}]{Hokama2014}
\bibinfo{author}{Hokama, P.}, \bibinfo{author}{San~Felice, M.~C.},
  \bibinfo{author}{Bracht, E.~C.}, \& \bibinfo{author}{Usberti, F.~L.}
  (\bibinfo{year}{2014}).
\newblock \bibinfo{title}{A heuristic approach for the stochastic steiner tree
  problem}.
\newblock In {\it \bibinfo{booktitle}{Proceedings of the 11th DIMACS
  Implementation Challenge in Collaboration with ICERM: Steiner Tree
  Problems}\/}.
\newblock \bibinfo{address}{Providence, RI, USA}.
\bibitem[{Hu et~al.(2014)Hu, Wen, Chua \& Li}]{Han-etal-2014}
\bibinfo{author}{Hu, H.}, \bibinfo{author}{Wen, Y.}, \bibinfo{author}{Chua,
  T.~S.}, \& \bibinfo{author}{Li, X.} (\bibinfo{year}{2014}).
\newblock \bibinfo{title}{Toward scalable systems for big data analytics: A
  technology tutorial}.
\newblock {\it \bibinfo{journal}{IEEE Access}\/},  {\it \bibinfo{volume}{2}\/},
  \bibinfo{pages}{652--687}.
\bibitem[{Johnson et~al.(1978)Johnson, Lenstra \&
  Rinnooy~Kan}]{JohnsonLenstraKanRinnooy1978}
\bibinfo{author}{Johnson, D.~S.}, \bibinfo{author}{Lenstra, J.~K.}, \&
  \bibinfo{author}{Rinnooy~Kan, A. H.~G.} (\bibinfo{year}{1978}).
\newblock \bibinfo{title}{The complexity of the network design problem}.
\newblock {\it \bibinfo{journal}{Networks}\/},  {\it \bibinfo{volume}{8}\/},
  \bibinfo{pages}{279--285}.
\bibitem[{Kara \& Onut(2010)}]{KaraOnut2010}
\bibinfo{author}{Kara, S.~S.}, \& \bibinfo{author}{Onut, S.}
  (\bibinfo{year}{2010}).
\newblock \bibinfo{title}{A two-stage stochastic and robust programming
  approach to strategic planning of a reverse supply network: The case of paper
  recycling}.
\newblock {\it \bibinfo{journal}{Expert Systems with Applications}\/},  {\it
  \bibinfo{volume}{37}\/}, \bibinfo{pages}{6129 -- 6137}.
\bibitem[{Krasko \& Rebennack(2017)}]{Krasko2017265}
\bibinfo{author}{Krasko, V.}, \& \bibinfo{author}{Rebennack, S.}
  (\bibinfo{year}{2017}).
\newblock \bibinfo{title}{Two-stage stochastic mixed-integer nonlinear
  programming model for post-wildfire debris flow hazard management: Mitigation
  and emergency evacuation}.
\newblock {\it \bibinfo{journal}{European Journal of Operational Research}\/},
  {\it \bibinfo{volume}{263}\/}, \bibinfo{pages}{265--282}.
\bibitem[{LEMON(2017)}]{Lemon2014}
\bibinfo{author}{LEMON} (\bibinfo{year}{2017}).
\newblock \bibinfo{title}{{LEMON} -- {L}ibrary for {E}fficient {M}odeling and
  {O}ptimization in {N}etworks}.
\newblock \bibinfo{howpublished}{Available at
  \url{http://lemon.cs.elte.hu/trac/lemon/}}.
\bibitem[{Lengauer(1990)}]{Lengauer1990}
\bibinfo{author}{Lengauer, T.} (\bibinfo{year}{1990}).
\newblock {\it \bibinfo{title}{Combinatorial Algorithms for Integrated Circuit
  Layout}\/}.
\newblock \bibinfo{address}{New York, NY, USA}: \bibinfo{publisher}{John Wiley
  \& Sons, Inc.}
\bibitem[{L{\o}kketangen \& Woodruff(1996)}]{LokketangenWoodruff1996}
\bibinfo{author}{L{\o}kketangen, A.}, \& \bibinfo{author}{Woodruff, D.~L.}
  (\bibinfo{year}{1996}).
\newblock \bibinfo{title}{Progressive hedging and tabu search applied to mixed
  integer (0,1) multistage stochastic programming}.
\newblock {\it \bibinfo{journal}{Journal of Heuristics}\/},  {\it
  \bibinfo{volume}{2}\/}, \bibinfo{pages}{111--128}.
\bibitem[{Magnanti \& Wong(1984)}]{MagnantiWong1984}
\bibinfo{author}{Magnanti, T.~L.}, \& \bibinfo{author}{Wong, R.~T.}
  (\bibinfo{year}{1984}).
\newblock \bibinfo{title}{Network design and transportation planning: Models
  and algorithms}.
\newblock {\it \bibinfo{journal}{Transportation Science}\/},  {\it
  \bibinfo{volume}{18(1)}\/}, \bibinfo{pages}{1--55}.
\bibitem[{McCarty \& Cohn(2018)}]{McCarty20181}
\bibinfo{author}{McCarty, L.}, \& \bibinfo{author}{Cohn, A.}
  (\bibinfo{year}{2018}).
\newblock \bibinfo{title}{Preemptive rerouting of airline passengers under
  uncertain delays}.
\newblock {\it \bibinfo{journal}{Computers and Operations Research}\/},  {\it
  \bibinfo{volume}{90}\/}, \bibinfo{pages}{1--11}.
\bibitem[{Parija et~al.(2004)Parija, Ahmed \& King}]{ParijaAhmedKing2004}
\bibinfo{author}{Parija, G.~R.}, \bibinfo{author}{Ahmed, S.}, \&
  \bibinfo{author}{King, A.~J.} (\bibinfo{year}{2004}).
\newblock \bibinfo{title}{On bridging the gap between stochastic integer
  programming and mip solver technologies}.
\newblock {\it \bibinfo{journal}{INFORMS J. on Computing}\/},  {\it
  \bibinfo{volume}{16}\/}, \bibinfo{pages}{73--83}.
\bibitem[{Restrepo et~al.(2017)Restrepo, Gendron \& Rousseau}]{Restrepo2017620}
\bibinfo{author}{Restrepo, M.}, \bibinfo{author}{Gendron, B.}, \&
  \bibinfo{author}{Rousseau, L.-M.} (\bibinfo{year}{2017}).
\newblock \bibinfo{title}{A two-stage stochastic programming approach for
  multi-activity tour scheduling}.
\newblock {\it \bibinfo{journal}{European Journal of Operational Research}\/},
  {\it \bibinfo{volume}{262}\/}, \bibinfo{pages}{620--635}.
\bibitem[{Rockafellar \& Wets(1991)}]{RockafellerWets1991}
\bibinfo{author}{Rockafellar, R.~T.}, \& \bibinfo{author}{Wets, R. J.-B.}
  (\bibinfo{year}{1991}).
\newblock \bibinfo{title}{Scenarios and policy aggregation in optimization
  under uncertainty}.
\newblock {\it \bibinfo{journal}{Mathematics of Operations Research}\/},  {\it
  \bibinfo{volume}{16}\/}, \bibinfo{pages}{119--147}.
\bibitem[{Shapiro et~al.(2014)Shapiro, Dentcheva \&
  Ruszczynski}]{shapiro2014lectures}
\bibinfo{author}{Shapiro, A.}, \bibinfo{author}{Dentcheva, D.}, \&
  \bibinfo{author}{Ruszczynski, A.} (\bibinfo{year}{2014}).
\newblock {\it \bibinfo{title}{Lectures on stochastic programming: modeling and
  theory}\/} volume~\bibinfo{volume}{16}.
\newblock \bibinfo{publisher}{SIAM}.
\bibitem[{Swamy \& Shmoys(2006)}]{Swamy2006}
\bibinfo{author}{Swamy, C.}, \& \bibinfo{author}{Shmoys, D.~B.}
  (\bibinfo{year}{2006}).
\newblock \bibinfo{title}{Approximation algorithms for 2-stage stochastic
  optimization problems}.
\newblock {\it \bibinfo{journal}{SIGACT News}\/},  {\it
  \bibinfo{volume}{37}\/}, \bibinfo{pages}{33--46}.
\bibitem[{Till et~al.(2007)Till, Sand, Urselmann \& Engell}]{TillEtAl2007}
\bibinfo{author}{Till, J.}, \bibinfo{author}{Sand, G.},
  \bibinfo{author}{Urselmann, M.}, \& \bibinfo{author}{Engell, S.}
  (\bibinfo{year}{2007}).
\newblock \bibinfo{title}{A hybrid evolutionary algorithm for solving two-stage
  stochastic integer programs in chemical batch scheduling}.
\newblock {\it \bibinfo{journal}{Computers and Chemical Engineering}\/},  {\it
  \bibinfo{volume}{31}\/}, \bibinfo{pages}{630 -- 647}.
\bibitem[{Tometzki \& Engell(2009)}]{TometzkiEngell2009}
\bibinfo{author}{Tometzki, T.}, \& \bibinfo{author}{Engell, S.}
  (\bibinfo{year}{2009}).
\newblock \bibinfo{title}{Hybrid evolutionary optimization of two-stage
  stochastic integer programming problems: An empirical investigation}.
\newblock {\it \bibinfo{journal}{Evolutionary Computation}\/},  {\it
  \bibinfo{volume}{17}\/}, \bibinfo{pages}{511--526}.
\bibitem[{Tometzki \& Engell(2011)}]{TometzkiEngell2011}
\bibinfo{author}{Tometzki, T.}, \& \bibinfo{author}{Engell, S.}
  (\bibinfo{year}{2011}).
\newblock \bibinfo{title}{Systematic initialization techniques for hybrid
  evolutionary algorithms for solving two-stage stochastic mixed-integer
  programs}.
\newblock {\it \bibinfo{journal}{IEEE Transactions on Evolutionary
  Computation}\/},  {\it \bibinfo{volume}{15}\/}, \bibinfo{pages}{196--214}.
\bibitem[{Vazirani(2003)}]{Va03}
\bibinfo{author}{Vazirani, V.~V.} (\bibinfo{year}{2003}).
\newblock {\it \bibinfo{title}{Approximation {A}lgorithms}\/}.
\newblock \bibinfo{address}{Germany}: \bibinfo{publisher}{Springer-Verlag
  Berlin Heidelberg}.
\bibitem[{Watson \& Woodruff(2010)}]{Watson2010}
\bibinfo{author}{Watson, J.-P.}, \& \bibinfo{author}{Woodruff, D.~L.}
  (\bibinfo{year}{2010}).
\newblock \bibinfo{title}{Progressive hedging innovations for a class of
  stochastic mixed-integer resource allocation problems}.
\newblock {\it \bibinfo{journal}{Computational Management Science}\/},  {\it
  \bibinfo{volume}{8}\/}, \bibinfo{pages}{355--370}.

\end{thebibliography}
